\documentclass[letterpaper, 10 pt, conference]{ieeeconf}  
\usepackage{setspace}

\IEEEoverridecommandlockouts
\usepackage{url}
\usepackage{balance}
\usepackage{amsmath}
\usepackage{amssymb,bm}
\usepackage{amsfonts}
\usepackage{enumerate}
\usepackage{tabulary}
\usepackage{multirow}
\usepackage{cite}
\usepackage{lipsum}  
\usepackage[pdftex]{graphicx}
\usepackage{epstopdf}
\usepackage[misc]{ifsym}
\usepackage{color}
\usepackage{xcolor}
\usepackage{xxcolor}
\usepackage{pgf}
\usepackage{pgfplots}
\usepackage{tikz}
\usepackage{float}
\usepackage{ifthen}
\usepackage{forloop}
\usepackage{listings}
\usepackage{lipsum}
\usepackage{booktabs}
\usepackage[ruled,linesnumbered]{algorithm2e}
\SetKwRepeat{Do}{do}{while}%

\usepackage{verbatim}
\usepackage{hyperref}
\usepackage{makecell}
\usepackage{xfrac}
\usepackage[font=footnotesize]{caption}
\usepackage{subcaption}
\usepackage{lipsum}

\DeclareMathOperator*{\argmin}{argmin}
\SetAlCapSkip{0.5em}
\newcommand*{\Scale}[2][4]{\scalebox{#1}{$#2$}}%

\title{\LARGE \bf
Finding the Right Place: Sensor Placement for UWB Time Difference of Arrival Localization in Cluttered Indoor Environments
}

\author{Wenda Zhao,  Abhishek Goudar, and Angela P. Schoellig
\thanks{The authors are with the \href{http://www.dynsyslab.org}{Dynamic Systems Lab}, Institute for Aerospace Studies, University of Toronto, Canada, and affiliated with the \href{https://vectorinstitute.ai/}{Vector Institute for Artificial Intelligence} in Toronto. 
        E-mails:
        {\tt \{firstname.lastname\}@utoronto.ca}}%
}

\begin{document}
\maketitle
\thispagestyle{empty}
\pagestyle{empty}

\begin{abstract}
Ultra-wideband (UWB) time difference of arrival (TDOA)-based localization has recently emerged as a promising indoor positioning solution. However, in cluttered environments, both the UWB radio positions and the obstacle-induced non-line-of-sight (NLOS) measurement biases  significantly impact the quality of the position estimate. Consequently, the placement of the UWB radios must be carefully designed to provide satisfactory localization accuracy for a region of interest. In this work, we propose a novel algorithm that optimizes the UWB radio positions for a pre-defined region of interest in the presence of obstacles. The mean-squared error (MSE) metric is used to formulate an optimization problem that balances the influence of the geometry of the radio positions and the NLOS effects. We further apply the proposed algorithm to compute a minimal number of UWB radios required for a desired localization accuracy and their corresponding positions. In a real-world cluttered environment, we show that the designed UWB radio placements provide $47\%$ and $76\%$ localization root-mean-squared error (RMSE) reduction in 2D and 3D experiments, respectively, when compared against trivial placements.

\end{abstract}

\setstretch{0.985}
\vspace{0.1em}
\section{Introduction}
\vspace{-0.2em}
\subsection{Motivation}
Accurate indoor localization is a crucial enabling technology for many robotics applications, from warehouse management to inspection and monitoring tasks. Over the last decade, ultra-wideband (UWB) radio technology has been shown to provide high-accuracy time-of-arrival (TOA) measurements, making it a promising indoor localization solution. In autonomous indoor robotics~\cite{kang2020ultra,hamer2018self}, the two main ranging schemes used for UWB localization are \textit{(i)}~two-way ranging (TWR) and \textit{(ii)}~ time difference of arrival (TDOA). In TWR, the UWB module mounted on the robot (also called tag) communicates with a fixed UWB radio (also called anchor) and acquires range measurements through two-way communication. In TDOA, UWB tags compute the difference between the arrival times of the radio packets from two anchors as TDOA measurements. Compared with TWR, TDOA does not require active two-way communication between an anchor and a tag, thus enabling localization of a large number of tags~\cite{hamer2018self}. We focus on UWB TDOA-based localization due to its scalability. 

Compared to other popular positioning approaches that rely on vision, ultra-sound, and laser ranging measurements, UWB-based localization, due to the obstacle-penetrating capability of UWB signals, does not require clear line-of-sight (LOS) conditions between UWB radios.
However, delayed and degraded radio signals caused by non-line-of-sight (NLOS) radio propagation lead to measurement biases and deteriorate localization accuracy. Apart from the measurement biases induced by NLOS, it is well known that the relative anchor-tag geometry has a significant influence on the localization performance, especially for TDOA-based positioning systems~\cite{Sathyan2010analysis,meng2016optimal}. Therefore, an optimized UWB anchor placement design balancing the effects of anchor-tag geometry and NLOS measurement biases is essential to guarantee reliable and accurate localization performance. 

In this work, we propose a novel sensor placement algorithm using the mean-squared error (MSE) metric to evaluate the TDOA-based localization performance, which considers both the influence of anchor-tag geometry and the NLOS effects. TDOA measurement models under various NLOS scenarios are constructed from experimental data. We formulate the anchor placement problem as an optimization problem and propose a block coordinate-wise minimization (BCM) algorithm to solve it numerically. We further apply the proposed algorithm to determine a minimal number of anchors and their corresponding positions to satisfy a given positioning accuracy requirement in a cluttered environment. The proposed algorithm can be applied to both 2D and 3D spaces. The effectiveness of our algorithm is validated through real-world experiments in both 2D and 3D scenarios. Our main contributions can be summarized as follows:
\begin{enumerate}
  \item We formulate the sensor placement optimization problem using the MSE metric, which considers both variance and bias of the position estimate.
  
  \item We construct UWB TDOA models under various NLOS conditions used in our sensor placement analysis. 
  
  \item We present a BCM-based algorithm to optimize the anchor positions in cluttered environments. Additionally, for localization system design, we find a minimal number of anchors and their corresponding positions to satisfy a given position accuracy requirement. 
 
  \item We validate the performance of the anchor placements through real-world experiments. We show that the anchor placements computed using the proposed algorithm yield significantly enhanced localization performance in both 2D and 3D scenarios.
\end{enumerate}

\subsection{Related Work}

Optimal sensor placement has most notably been studied for the Global Positioning System (GPS)~\cite{enge1994global}. Geometric Dilution of Precision (GDOP) is a classical performance metric widely used to evaluate the quality of a GPS satellite configuration~\cite{sharp2009gdop}. The Cram\'{e}r-Rao lower bound (CRLB)~\cite{chaffee1994gdop} is a more general performance metric, which considers the statistical properties of measurements for evaluating estimation performance. Most existing works focus on the optimal sensor placement w.r.t. a single location of interest, also called target point~\cite{meng2016optimal}. In this context, the TDOA optimal sensor geometry problem for both static and dynamic sensor setups was considered in~\cite{wang2018optimal}, and extended to 3D in~\cite{xu2019optimal}.
%
In its general form, the sensor placement problem for multiple target points has been shown to be NP-hard~\cite{tekdas2010sensor}. Pioneering work in \cite{jourdan2008optimal} addresses the general sensor placement problem with range measurements using a numerical algorithm based on the Position Error Bound (PEB) metric~\cite{jourdan2008position}. Our work differs in terms of the ranging scheme (TDOA), the NLOS model, and the performance metric considered. Existing sensor placement algorithms often rely on the assumption that the position estimation results at target points are unbiased. However, with obstacle-induced measurement biases, the position estimate typically does not remain unbiased. Consequently, both variance and bias of the position estimate must be considered during sensor placement design.

In this work, we use the MSE metric to formulate the sensor placement optimization, which explicitly considers both variance and bias of the estimation. We demonstrate the effectiveness of the proposed method through real-world 2D and 3D experiments. To the best of our knowledge, this is the first work to consider both variance and bias of the position estimate in sensor placement design. 

\vspace{0.5em}
\section{Problem Statement}
\label{sec:problem}
\subsection{Preliminaries and Notation}
We study the sensor placement problem for a decentralized TDOA localization system, which computes the TDOA measurements between anchor pairs and is not impacted by communication constraints and single-anchor failures~\cite{ennasr2016distributed}. The set of $m$ UWB anchors are divided into disjoint pairs $\Gamma = \{(1,2),\cdots, (m-1,m)\}$ and are assumed to be fixed in an indoor space $\mathcal{P}\in \mathbb{R}^{n}$ with $n=\{2,3\}$ indicating the dimension. To facilitate our analysis, we define a vector $\bm{a}=[\bm{a}_1^T, \bm{a}_2^T, \cdots, \bm{a}_m^T]^T \in \mathbb{R}^{n \cdot m}$ that contains all anchor positions. The position of the UWB tag is $\mathbf{p} \in \mathcal{P}$. We consider a realistic indoor environment and assume stationary obstacles in $\mathcal{P}$ with known positions. The obstacles are categorized into blocking $\mathcal{O}_b$, metal $\mathcal{O}_m$, and non-metal $\mathcal{O}_{nm}$ (e.g., wood or plastic) obstacles for UWB NLOS error modeling. More formally, we define a set $\mathcal{O} = \mathcal{O}_b\bigcup\mathcal{O}_m\bigcup\mathcal{O}_{nm} \subset \mathcal{P}$ to denote the space covered by the obstacles of each category. Bounding boxes are used to represent the obstacles and NLOS conditions are detected through ray tracing. We define the region of interest as $\Phi \subset \left(\mathcal{P} \cap \mathcal{O}'\right)$ with $\mathbf{p} \in \Phi$, where $\mathcal{O}'$ is the complement of set $\mathcal{O}$. The localization performance in $\Phi$ is optimized through the anchor positions $\bm{a}$.

\subsection{The Optimal Sensor Placement Problem}
Our problem is described as follows. For a finite indoor space $\mathcal{P}$ cluttered by obstacles $\mathcal{O} \subset \mathcal{P}$, we aim to optimize the anchor positions $\bm{a}$ for a decentralized UWB TDOA-based localization system w.r.t. a pre-defined region of interest $\Phi \subset \left(\mathcal{P} \cap \mathcal{O}'\right)$. To facilitate our analysis, we evenly sample $N$ points $\mathbf{p}_i \in \Phi, i=1\cdots N$, for evaluation to represent the localization performance in $\Phi$.
%
%
An error-free TDOA measurement at position $\mathbf{p}$ computes the range difference between $\{\bm{a}_i,\mathbf{p}\}$ and $\{\bm{a}_j,\mathbf{p}\}$:
\begin{equation}
\label{eq:ideal-tdoa-model}
\bar{d}_{ij}(\mathbf{p}, \bm{a}_i,\bm{a}_j) = \|\mathbf{p} - \bm{a}_j\| - \|\mathbf{p} - \bm{a}_i\|, (i,j)\in\Gamma, 
\end{equation}
where $\|\cdot\|$ indicates the $\ell_2$ norm. Considering measurement noise $\epsilon_{ij}$ and obstacle-induced NLOS bias $b_{ij,\textrm{nlos}}$, the noisy TDOA measurements are modeled as 
\begin{equation}
\label{eq:noisy-tdoa-model}
   d_{ij}(\mathbf{p},\bm{a}_i,\bm{a}_j)  = \bar{d}_{ij}(\mathbf{p},\bm{a}_i,\bm{a}_j) +
   \varepsilon_{ij} + b_{ij,\textrm{nlos}}, 
\end{equation}
and assumed to be independent with $\varepsilon_{ij} \sim \mathcal{N}(0, \sigma^2)$ being a zero-mean Gaussian distribution with variance $\sigma^2$, common to all TDOA measurements. 
For brevity, we drop the functional dependence and denote the UWB TDOA measurements as $\mathbf{d}=[d_{12},\cdots, d_{(m-1)m}]^T\in  \mathbb{R}^{Q}$ with $Q = m/2$ indicating the number of anchor pairs. Stacking the measurements into vector form, we have 
\begin{equation}
\label{eq:tdoa-model-vec}
\mathbf{d} = \bar{\mathbf{d}}(\mathbf{p}, \bm{a}) + \bm{\varepsilon} + \mathbf{b}_{\textrm{nlos}}
\end{equation}
with probability density function $f(\mathbf{d}|\mathbf{p},\bm{a})$, where $\bm{\varepsilon}$ and $\mathbf{b}_{\textrm{nlos}}$ are the vector forms of the measurement noise and the NLOS bias modelled as random variables. 

To evaluate the localization performance of a given anchor placement $\bm{a}$ in a cluttered environment, we first derive a general localization performance metric at one target point $\mathbf{p}$, assuming $f(\mathbf{d}|\mathbf{p},\bm{a})$ is known. We then derive a model for $f(\mathbf{d}|\mathbf{p},\bm{a})$, which depends on the NLOS conditions induced by the obstacles $\mathcal{O} \subset \mathcal{P}$. With the general performance metric and the NLOS models, the third goal is to develop an algorithm that optimizes the anchor locations $\bm{a}$ for a pre-defined region of interest $\Phi$.

\section{Optimal Sensor Placement Analysis}
\label{sec:estimation_theory}
The placement of UWB anchors is critical to ensure good localization accuracy. A classic performance metric for the sensor placement problem is the CRLB, which characterizes the minimal achievable variance of an unbiased estimator. However, considering the obstacle-induced measurement biases, the position estimate typically does not remain unbiased and the CLRB is no longer a suitable performance benchmark. To analyze sensor placement in cluttered environments, we first derive the mean-squared error (MSE) performance metric for a general UWB TDOA positioning system, considering both the variance and the bias of the position estimate. We then formulate the anchor placement optimization problem. 

\subsection{Mean-Squared Error Metric}
\label{subsec:mse}
For a decentralized UWB TDOA localization system with known anchor positions $\bm{a}$, the tag position $\mathbf{p}$ is estimated from TDOA measurements $\mathbf{d}$ with probability density function $f(\mathbf{d}|\mathbf{p},\bm{a})$. The estimated tag position $\hat{\mathbf{p}}$ is characterized by its bias $\textrm{Bias}(\hat{\mathbf{p}})$ and covariance matrix $\textrm{Cov}(\hat{\mathbf{p}})$. In this work, we rely on the MSE of the position estimate $\hat{\mathbf{p}}$ as the performance metric, which evaluates the error between the estimate $\hat{\mathbf{p}}$ and the true value $\mathbf{p}$~\cite{eldar2008rethinking}. The MSE of an estimate $\hat{\mathbf{p}}$ of $\mathbf{p}$ can be decomposed as 
\begin{equation}
    \textrm{MSE}(\hat{\mathbf{p}}) = \mathbb{E}\{\|\hat{\mathbf{p}} - \mathbf{p}\|^2\} = \textrm{Tr}\left(\textrm{Cov}(\hat{\mathbf{p}})\right) + \|\textrm{Bias}(\hat{\mathbf{p}})\|^2,
\end{equation}
where $\textrm{Tr}(\cdot)$ is the trace operator. We assume $\textrm{Bias}(\hat{\mathbf{p}})$ is induced by TDOA measurement noise and obstacle-induced NLOS bias. 

To evaluate the $\textrm{MSE}(\hat{\mathbf{p}})$ for a given anchor placement $\bm{a}$, we first compute $\textrm{Bias}(\hat{\mathbf{p}})$ approximately through linearization. Assuming that $\hat{\mathbf{p}}$ is sufficiently close to $\mathbf{p}$, we linearize the error-free measurement model $\bar{\mathbf{d}}(\mathbf{p}, \bm{a})$ around $\mathbf{p}$, ignore the higher-order terms, and obtain a relationship between the estimation error and the corresponding measurement error: $\Delta \mathbf{d} = \mathbf{J}(\mathbf{p}, \bm{a})(\hat{\mathbf{p}} - \mathbf{p})$, where $\mathbf{J}(\mathbf{p}, \bm{a})$ is the Jacobian matrix of $\bar{\mathbf{d}}(\mathbf{p}, \bm{a})$ w.r.t. $\mathbf{p}$. From \eqref{eq:tdoa-model-vec}, we see that $\Delta \mathbf{d} = \mathbf{b}_{\textrm{nlos}} + \bm{\varepsilon}$ and obtain:
\begin{equation}
    \hat{\mathbf{p}} - \mathbf{p} = \mathbf{J}(\mathbf{p}, \bm{a})^{\dagger}(\mathbf{b}_{\textrm{nlos}} + \bm{\varepsilon}),
\end{equation}
where $\mathbf{J}(\mathbf{p}, \bm{a})^{\dagger} = (\mathbf{J}(\mathbf{p}, \bm{a})^T \mathbf{J}(\mathbf{p}, \bm{a}))^{-1}\mathbf{J}(\mathbf{p}, \bm{a})^T$ is the pseudoinverse matrix. Therefore, the estimation bias can be computed as 
\begin{equation}
\label{eq:bias-est}
    \textrm{Bias}(\hat{\mathbf{p}}) = \mathbb{E}(\hat{\mathbf{p}}) - \mathbf{p}  = \mathbf{J}(\mathbf{p},\bm{a})^{\dagger} \mathbb{E}(\mathbf{b}_{\textrm{nlos}}) := \bm{\beta}(\mathbf{p}),
\end{equation}
which is a function of $\mathbf{p}$. Note, we assume the anchor placement $\bm{a}$ is given and drop the functional dependency.

Based on the uniform Cram\'{e}r-Rao bound~\cite{eldar2008rethinking}, the covariance of the estimate $\hat{\mathbf{p}}$ of $\mathbf{p}$ with the estimation bias $\bm{\beta}(\mathbf{p})$ is lower bounded by 
\begin{equation}
    \label{eq:CRB}
    \textrm{Cov}(\hat{\mathbf{p}}) \succeq (\mathbf{I}+\mathbf{D}(\mathbf{p}))\mathcal{I}^{-1}(\mathbf{p})(\mathbf{I}+\mathbf{D}(\mathbf{p}))^T,
\end{equation}
where $\succeq$ denotes the generalized inequality~\cite{Boyd1994matrix} and $\mathcal{I}(\mathbf{p})$ is the Fisher Information Matrix (FIM) defined by 
\begin{equation}
    \mathcal{I}(\mathbf{p}) = \mathbb{E}\left\{ \left[\frac{\partial \log f(\mathbf{d}|\mathbf{p},\bm{a})}{\partial \mathbf{p}}\right]\left[\frac{\partial \log f(\mathbf{d}| \mathbf{p},\bm{a})}{\partial \mathbf{p}}\right]^{T}\right\},
\end{equation}
and $\mathbf{D}(\mathbf{p}) = \partial \bm{\beta}(\mathbf{p})/ \partial \mathbf{p}$ is the bias gradient matrix. Since the TDOA measurements are assumed to be independent, under minor assumptions~\cite{zegers2015fisher}, the FIM can be computed as a matrix with the $(u,v)$th element being
\begin{equation}
\label{eq:fim}
\Scale[0.97]{
\begin{split}
    \Big[\mathcal{I}(\mathbf{p})\Big]_{u,v} 
    & = - \mathbb{E}\left[ \frac{\partial^2 \log \left(\prod_{(i,j)\in\Gamma} f(d_{ij}|\mathbf{p},\bm{a})\right)}{\partial p_u \partial p_v} \right]  \\
    & = -\sum_{(i,j)\in \Gamma} \mathbb{E}\left[\frac{\partial^2 \log f(d_{ij}|\mathbf{p},\bm{a})}{\partial p_u \partial p_v}\right],
\end{split}
}
\end{equation}
where $p_u$ and $p_v$ are the $u$th and $v$th element of $\mathbf{p}$. Therefore, we can compute the lower bound of the $\textrm{MSE}(\hat{\mathbf{p}})$ based on Equation~\eqref{eq:bias-est},~\eqref{eq:CRB}, and~\eqref{eq:fim} as
\begin{equation}
\label{eq:mse_benchmark}
\Scale[0.95]{
    M(\mathbf{p}) = \textrm{Tr}\left((\mathbf{I}+\mathbf{D}(\mathbf{p}))\mathcal{I}^{-1}(\mathbf{p})(\mathbf{I}+\mathbf{D}(\mathbf{p}))^T \right) + \|\bm{\beta}(\mathbf{p})\|^2.
    }
\end{equation}
We compute $M(\mathbf{p})$ to evaluate the localization performance at the target point $\mathbf{p}$. The proposed MSE metric applies to any probability density functions for which the FIM exists.

\subsection{Sensor Placement Optimization}
\label{sec:spo}
In realistic scenarios, UWB anchors have limited operating range, denoted here by $r_{\textrm{op}}$. Moreover, UWB radio signals can be completely blocked by blocking obstacles $\mathcal{O}_b$. To account for such phenomena during anchor placement design, we assume UWB measurements are unavailable when anchors are out of operating range or any two UWB radios, one anchor and one tag or two anchors, are blocked by $\mathcal{O}_b$. With $r_{\textrm{max}} = \max \{\|\mathbf{p} - \bm{a}_i\|, \|\mathbf{p} - \bm{a}_j\|, \|\bm{a}_i - \bm{a}_j\|\}$ being the largest distance among UWB radios, we introduce the following weight functions: 
\begin{equation}
\label{eq:weight}
\Scale[0.95]{
    w^{r}_{ij}(\mathbf{p},\bm{a}) =  \begin{cases}
                                  1,       & r_{\textrm{max}} \leq r_{\textrm{op}}   \\
                                  0,       & \textrm{otherwise}
                                \end{cases},
    w^{b}_{ij}(\mathbf{p},\bm{a}) =  \begin{cases}
                                  0,       & \mathcal{O}_b ~\textrm{NLOS} \\
                                  1,       & \textrm{otherwise}.
                                \end{cases}
            }
\end{equation}
For brevity, we drop the functional dependencies and define a general weight $w_{ij} = w^{r}_{ij}w^{b}_{ij}$. We adapt $M(\mathbf{p})$ in~\eqref{eq:mse_benchmark} to account for the range limitation and signal blockage by defining a matrix $\mathbf{W}$ with $w_{ij}$ on the diagonal entries and introducing the intermediate variables:
\begin{equation}
\label{eq:weight_fim_bias}
\Scale[0.97]{
\begin{split}
\left[\tilde{\mathcal{I}}(\mathbf{p})\right]_{u,v} 
 & = -\sum_{(i,j)\in \Gamma} w_{ij} \mathbb{E}\left[\frac{\partial^2 \log f(d_{ij}| \mathbf{p}, \bm{a})}{\partial p_u \partial p_v}\right], \\
\tilde{\mathbf{J}}(\mathbf{p}, \bm{a}) &= \mathbf{W}\mathbf{J}(\mathbf{p}, \bm{a}), ~~ \mathbb{E}(\tilde{\mathbf{b}}_{\textrm{nlos}}) = \mathbf{W}\mathbb{E}(\mathbf{b}_{\textrm{nlos}}).
\end{split}
}
\end{equation}
The weighted estimation bias $\tilde{\bm{\beta}}(\mathbf{p})$ and bias gradient matrix $\tilde{\mathbf{D}}(\mathbf{p})$ can be obtained accordingly. We compute the weighted MSE metric $\tilde{M}(\mathbf{p})$ using Equation~\eqref{eq:mse_benchmark} with $\tilde{\mathbf{D}}(\mathbf{p})$, $\tilde{\bm{\beta}}(\mathbf{p})$, and $\tilde{\mathcal{I}}(\mathbf{p})$. One can show that $\tilde{M}(\mathbf{p})$ is equivalent to discarding the blocked and out-of-range UWB measurements in the original computation of $M(\mathbf{p})$.

For sensor placement analysis, we denote the weighted MSE error metric as $\tilde{M}(\mathbf{p}, \bm{a})$ and reintroduce its dependency on $\bm{a}$. To represent the localization performance within an entire region $\Phi$, we evaluate the weighted MSE metric at the $N$ sample points $\mathbf{p}_i \in \Phi, i=1\cdots N$, and compute the average root-mean-squared error (RMSE) 
\begin{equation}
\label{eq:mse_metric}
    \mathcal{M}(\bm{a}) = \frac{1}{N}\sum_{i=1}^{N} \sqrt{\tilde{M}(\mathbf{p}_i, \bm{a})}
\end{equation}
as the performance metric. We propose that the optimal UWB anchor placement w.r.t. region $\Phi$ corresponds to the anchor placement leading to the smallest value $\mathcal{M}(\bm{a})$. As the anchors cannot be placed inside obstacles, we define a set $\mathcal{A}$ containing all possible anchor configurations with $\bm{a}\in \mathcal{A}$. We aim to find the optimal placement of UWB anchors $\bm{a}^{\star}$ that minimizes $\mathcal{M}(\bm{a})$:
\begin{equation}
\label{eq:optimization_problem}
    \bm{a}^{\star} = \argmin_{\bm{a}\in \mathcal{A}} \mathcal{M}(\bm{a}).
\end{equation}

\section{UWB TDOA Measurement Model}
\label{sec:tdoa_mm}
To perform sensor placement analysis, we model UWB TDOA measurements $f(\mathbf{d} | \mathbf{p},\bm{a})$ by constructing representative measurement error distributions  $f(\Delta \mathbf{d} | \mathbf{p},\bm{a})$ under various NLOS scenarios.
\begin{figure*}[!t]
\includegraphics[width=.96\textwidth]{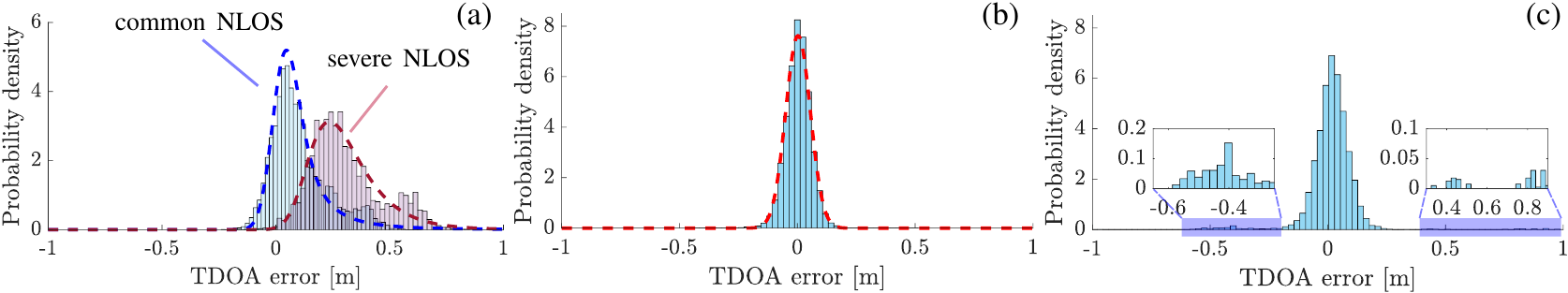}
\caption{Histograms of measurement errors induced by~(a) common and severe NLOS between anchor $\bm{a}_j$ and the tag \big(NLOS event $(L_i, \bar{L}_j, L_{ij})$\big), (b)~common NLOS between the two anchors \big(NLOS event $(L_i, L_j, \bar{L}_{ij})$\big), and (c)~severe NLOS event $(L_i, L_j, \bar{L}_{ij})$. The modeled probability density functions of the measurement errors  are indicated as dashed lines in~(a) and~(b), respectively. The severe NLOS event $(L_i, L_j, \bar{L}_{ij})$ induces measurement errors with a long-tailed distribution, see~(c). }
\label{fig:nlos-model}
\end{figure*}
Although UWB signals can pass through obstacles, NLOS radio propagation causes delayed or degraded signals and makes TOA detection algorithms prone to error, leading to inaccurate measurements. As TDOA-based localization involves three UWB radios, the NLOS scenarios are inherently complicated. In this section, we first model measurement biases under NLOS between one anchor and one tag and then between two anchors using experimental data. We later extend the model to general NLOS conditions.

\subsection{General UWB TDOA Measurement Model}
\label{subsec:general_tdoa_mm}
The UWB TDOA NLOS measurement errors are identified using Bitcraze's Loco Positioning System (LPS), while the modeling procedure can be generalized to other UWB platforms. As explained in Section~\ref{sec:spo}, we discard the measurements when UWB radios are blocked by $\mathcal{O}_b$ and focus on modeling the measurement errors induced by $\mathcal{O}_m$ and $\mathcal{O}_{nm}$. According to the TDOA principles explained in our dataset paper~\cite{zhao22uwb_dataset}, UWB TDOA measurements are computed based on radio packets received by the tag and exchanged between two anchors. Consequently, NLOS measurement errors could be induced by NLOS between one anchor and one tag, between the anchors, or combinations of the two.

Following the convention in~\cite{prorok2012online}, we define the LOS events between the tag and anchor $\bm{a}_i$, between the tag and anchor $\bm{a}_j$, and between the anchors as $\{L_i, L_j, L_{ij}\}$. The corresponding NLOS events are indicated as $\{\bar{L}_i, \bar{L}_j, \bar{L}_{ij}\}$. Given the TDOA model in~\eqref{eq:noisy-tdoa-model}, the measurement error $\Delta d_{ij}$ is composed of the measurement noise $\varepsilon_{ij}\sim \mathcal{N}(0, \sigma^2)$ and the NLOS bias $b_{ij,\textrm{nlos}}$. The variance $\sigma^2$ is obtained from the LOS experiment. Based on the TDOA principles~\cite{zhao22uwb_dataset}, we model the NLOS bias $b_{ij,\textrm{nlos}}$ as: 
\begin{equation}
\label{eq:general_tdoa_mm}
b_{ij,\textrm{nlos}}  = Y_i b_i + Y_j b_j +  Y_{ij} b_{ij}, (i,j) \in \Gamma,
\end{equation}
where $\{b_{i}, b_{j}, b_{ij}\}$ are random variables that represent the measurement errors induced by NLOS events $\{\bar{L}_i, \bar{L}_j, \bar{L}_{ij}\}$, respectively. Note, this NLOS bias model may vary for a different UWB TDOA platform. The independent random variables $Y_i$, $Y_j$, and $Y_{ij}$ follow Bernoulli distributions, which take the value $1$ with the probability of the corresponding NLOS events. Given positions of obstacles, anchors $\bm{a}$, and the tag $\mathbf{p}$, NLOS conditions can be detected through ray tracing. In practice, users can enlarge the sizes of obstacles to consider a conservative scenario. In this work, $Y_i$, $Y_j$, and $Y_{ij}$ take the value $1$ when the corresponding NLOS event is detected and $0$ otherwise. Large erroneous measurements caused by multi-path propagation are assumed to be rejected as outliers.

To model $b_i$, $b_j$, and $b_{ij}$, we conducted extensive experiments for the NLOS events $(L_i,\bar{L}_j,L_{ij})$ and $(L_i,L_j,\bar{L}_{ij})$ using obstacles of different types of materials, including cardboard, metal, wood, plastic, and foam. Readers are referred to our dataset paper~\cite{zhao22uwb_dataset} for all the details. Through the NLOS experiments, we observe that NLOS events induced by foam, cardboard or plastic obstacles have little influence on the UWB TDOA measurements. In contrast, both wooden and metal obstacles cause measurement errors with the latter inducing a considerably larger effect. To obtain a simple and practical UWB NLOS model, we introduce two different NLOS categories: \textit{(i) common NLOS} represents NLOS induced by $\mathcal{O}_{nm}$ (including foam, plastic, wood, etc.) and is conservatively modeled using the measurement errors induced by wooden obstacles; \textit{(ii) severe NLOS} represents NLOS induced by $\mathcal{O}_m$ or both $\mathcal{O}_m$ and  $\mathcal{O}_{nm}$. This two-category model is useful in practice, where the exact material of obstacles is often not known. 

We denote the probability density function of $\varepsilon_{ij}$ by $f_{\mathcal{N}}$. Since noise is always present, $f_{\mathcal{N}}$ will appear in the probability densities for all events. For the NLOS event $(L_i,\bar{L}_j,L_{ij})$, we summarize the measurement errors of common and severe NLOS conditions in Figure~\ref{fig:nlos-model}a. Despite the complexity of NLOS error patterns, we model the measurement bias $b_j$ induced by common and severe NLOS event $\bar{L}_j$ as log-normal distributions $f^{c}_{ln\mathcal{N},j}$ and $f^{s}_{ln\mathcal{N},j}$ to characterize the experimental data. For ease of notation, we drop the NLOS category superscripts $c$ and $s$ and indicate the bias distribution as $f_{ln\mathcal{N},j}$ for $(L_i,\bar{L}_j,L_{ij})$.
Analogously, for $(\bar{L}_i,L_j,L_{ij})$ we model the measurement bias $b_i$ as negative log-normal distributions $f^{-}_{ln\mathcal{N},i}(\tau) = f_{ln\mathcal{N},i}(-\tau)$ with $\tau \in\mathbb{R}$. 
The probability density of the measurement error in events $(L_i,\bar{L}_j,L_{ij})$ and $(\bar{L}_i,L_j,L_{ij})$ can be written as 
\begin{equation}
\begin{split}
    f(\Delta d_{ij} | L_i, \bar{L}_j, L_{ij}) &= (f_{ln\mathcal{N},j}*f_{\mathcal{N}})(\Delta d_{ij})\\
    f(\Delta d_{ij} | \bar{L}_i, L_j, L_{ij}) &= (f^{-}_{ln\mathcal{N},i}*f_{\mathcal{N}})(\Delta d_{ij}),
\end{split}
\label{eq:nlos_event_bi_bj}
\end{equation}
where the $*$ operator indicates the convolution of two probability distributions.
Since we have previously identified $f_{\mathcal{N}}$ from LOS experiments, we can now obtain the parameters of $f_{ln\mathcal{N},j}$ and $f^{-}_{ln\mathcal{N},i}$ through maximum likelihood estimation.

For the NLOS event $(L_i,L_j,\bar{L}_{ij})$, we observe that the measurement errors remain zero-mean for both common and severe NLOS (see Figure~\ref{fig:nlos-model}b-c, respectively). This phenomenon can be explained by how radio packets travel between the anchors~\cite{zhao22uwb_dataset}. Unlike the common NLOS errors which can be modeled as a Gaussian distribution $f^{c}_{ij}$, the severe NLOS errors result in a long-tailed distribution $f^{s}_{ij}$. For brevity, we indicate the probability density of $b_{ij}$ as $f_{ij}$ by dropping the category superscripts and model the measurement errors under $(L_i, L_j, \bar{L}_{ij})$ as
\begin{equation}
    f(\Delta d_{ij} | L_i, L_j, \bar{L}_{ij}) = (f_{ij}*f_{\mathcal{N}})(\Delta d_{ij}).
    \label{eq:nlos_event_b_ij}
\end{equation}

With $f_{ln\mathcal{N},j}$, $f^{-}_{ln\mathcal{N},i}$, and $f_{ij}$, we can derive the probability density function $f_{b_{ij},\textrm{nlos}}$ for $b_{ij,\textrm{nlos}}$ in Equation~\eqref{eq:general_tdoa_mm} for different NLOS conditions through convolution, and obtain the measurement error model
\begin{equation}
    f(\Delta d_{ij}) = (f_{b_{ij},\textrm{nlos}} * f_{\mathcal{N}})(\Delta d_{ij}).
\end{equation}

\subsection{Efficient TDOA Measurement Model}
\label{subsec:efficient-tdoa-model}
Although the FIM required for the proposed MSE metric in Equation~\eqref{eq:mse_benchmark} can be computed numerically~\cite{spall2005monte} with the models constructed above, directly using them for anchor placement analysis is computationally expensive. Hence, for sensor placement analysis, we approximately model the NLOS measurement errors as closed-form Gaussian distributions which allow us to analytically compute the FIM (see the supplementary material
\footnote{\hypertarget{fn:supplementary}{\url{http://tiny.cc/supplementary_doc}}}).

\begin{figure}[tb]
  \begin{tikzpicture}
      \centering
      \node[inner sep=0pt, opacity=1] (anchor) at (0.0,0.0){\includegraphics[width=.46\textwidth]{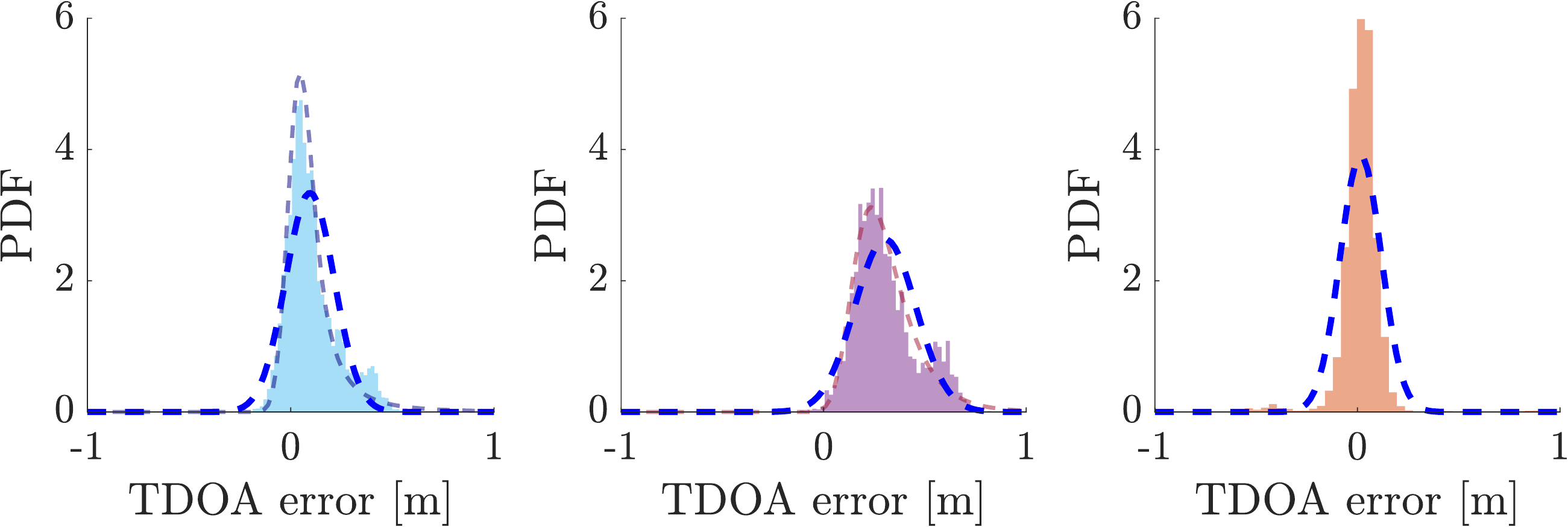}};
   \node[text width=2.5cm, black] at (-0.7, 1.25) {(a)};
   \node[text width=2.5cm, black] at (2.0, 1.25) {(b)};
   \node[text width=2.5cm, black] at (4.9, 1.25) {(c)};
   \end{tikzpicture}
  \caption{Approximated probability density functions (Gaussian distributions indicated by the dashed lines) in~(a) the common NLOS event $(L_i,\bar{L}_j, L_{ij})$, (b)~the severe NLOS event $(L_i, \bar{L}_j, L_{ij})$, and (c)~the severe NLOS event $(L_i,L_j,\bar{L}_{ij})$ are shown together with the experimental data (histogram).}
  \label{fig:variation_app}
\end{figure}
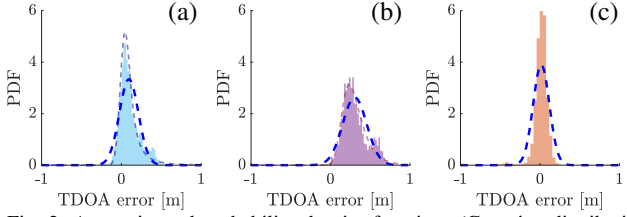

For the NLOS events $(L_i,\bar{L}_j,L_{ij})$, $(\bar{L}_i,L_j,L_{ij})$, and $(\bar{L}_i, \bar{L}_j,L_{ij})$, we approximate the probability density function of the NLOS bias by finding the closest Gaussian distributions, which we obtain through minimizing the Kullback-Leibler (KL) divergence~\cite{kullback1997information}. As numerically validated in~\cite{prorok2012online}, the Gaussian approximation of $f_{ln\mathcal{N},j} * f^{-}_{ln\mathcal{N},i}$ in $(\bar{L}_i, \bar{L}_j,L_{ij})$ shows a good match of probability densities.
For the long-tail distribution in the severe NLOS event $(L_i, L_j,\bar{L}_{ij})$, we construct a Gaussian distribution with the mean and variance computed from experimental data to represent the uncertainty of the NLOS biases. 
The approximated Gaussian distributions in common and severe NLOS events $(L_i,\bar{L}_j,L_{ij})$ and severe NLOS event $(L_i,L_j,\bar{L}_{ij})$ are shown in Figure~\ref{fig:variation_app} with the corresponding experimental data.

Since the NLOS error models for the $(\bar{L}_i,L_j,L_{ij})$, $(L_i,\bar{L}_j,L_{ij})$, $(\bar{L}_i,\bar{L}_j,L_{ij})$, and  $(L_i,L_j,\bar{L}_{ij})$ events are approximated as Gaussian distributions, all other NLOS error models can be derived through convolutions of Gaussian distributions. Considering three possible conditions (LOS, common NLOS, and severe NLOS) between either two UWB radios, we have in total $27$ TDOA models for anchor placement analysis. 

\begin{algorithm}[t]
\footnotesize
  \SetKwInput{KwData}{Input    ~}
  \SetKwInput{KwResult}{Output}
  \KwData{Number of anchor pairs $Q = m/2$, indoor space $\mathcal{P}$, possible anchor placement set $\mathcal{A}$, locations and materials of obstacles $\mathcal{O}$, pre-defined region of interest $\Phi \subset \left(\mathcal{P} \cap \mathcal{O}'\right)$, initial anchor positions $\Scale[1.0]{\bm{a}^{1}=\{\bm{a}^{1}_1,\cdots,\bm{a}^{1}_{m}\} \in \mathcal{A}}$, $k = 1$, and the maximum iteration max$_{\textrm{iter}}$.}
  \KwResult{Optimized placement $\Scale[1.0]{\bm{a}^{\star} \in \mathcal{A}}$.} 
  Select $N$ sample points $\mathbf{p}=\{\mathbf{p}_1,\cdots, \mathbf{p}_N\} \in \Phi$\; 
  \While{$k<\textrm{max}_{iter}$}{
    \For{$q\gets1$ \KwTo $Q$}{
        Select the $q$-th anchor pair $\{\bm{a}^{k}_i,\bm{a}^{k}_j\}_q$ \;
        Optimize $\{\bm{a}^{k\ast}_i,\bm{a}^{k\ast}_j\}_q$ to minimize the average RMSE over $\mathbf{p}\in \Phi$ in Equation \eqref{eq:mse_metric} \;
        Update the anchor pair $\{\bm{a}^{k}_i,\bm{a}^{k}_j\}_q \gets \{\bm{a}^{k\ast}_i,\bm{a}^{k\ast}_j\}_q$\;
    }
    $k \gets k+1$
  }
  \Return Optimized placement $\bm{a}^{\star}$ 
  \caption{\small BCM-based anchor placement algorithm}
\end{algorithm}
\section{Algorithm}
\label{sec:algorithm}
Given a general environment with known obstacles, applying the NLOS models constructed in Section~\ref{sec:tdoa_mm} to solve Equation~\eqref{eq:optimization_problem} is not trivial. %
In this work, we propose a block coordinate-wise minimization (BCM) algorithm~\cite{tseng2001convergence} to optimize Equation~\eqref{eq:optimization_problem}, which tends to be more efficient and scalable than solving the problem directly using nonlinear optimization solvers.
In the proposed BCM algorithm (see Algorithm~1), each anchor pair $\{\bm{a}_i, \bm{a}_j\}, (i,j)\in\Gamma,$ is treated as one coordinate block. At each iteration, we sequentially select one of the coordinate blocks and minimize the average RMSE $\mathcal{M}(\bm{a})$ w.r.t. the selected block while the other coordinate blocks are fixed. We use the \texttt{MultiStart}~\cite{matlab} solver provided by Matlab with $50$ random initial conditions for each optimization step (line~5 in Algorithm~1). 
In practice, operating the algorithm with a fixed number of iterations (max$_{\textrm{iter}}=5$), the algorithm typically converges to satisfactory placement solutions. 
Like for any general nonlinear optimization problem, we do not have guarantees for finding the global optimal solution, but our numerical solutions demonstrate significantly better performance in both simulation and experiments compared to trivial or intuitive anchor placements (shown in Section~\ref{sec:sim-exp}).

We further use the sensor placement optimization for localization system design. Given the obstacle positions, a pre-defined region of interest $\Phi$, and required localization accuracy $\mathcal{M}_r$, we propose an iterative algorithm based on Algorithm~1 to find a minimal number of anchors required to meet the accuracy $\mathcal{M}_r$ and their corresponding positions. Considering complicated indoor environments with solid walls and other obstacles, we constrain the anchor positions to the boundary of the indoor space. In each iteration, we optimize the anchor placement and add one extra pair of anchors if the desired localization accuracy is not satisfied. The positions of the extra anchor pair are initialised randomly on the boundary of the space. The algorithm terminates when the required accuracy is satisfied or no solution can be found with the maximum allowable number of anchors. The detailed localization system design algorithm is presented in the supplementary material\hyperlink{fn:supplementary}{$^1$} (Algorithm~2).

\section{Simulation and Experimental Results}
\label{sec:sim-exp}
In this section, we present simulation and experimental results of the proposed sensor placement algorithm. The algorithm is implemented in Matlab and we use Bitcraze's Loco Positioning system (LPS) for experimental validation. We first demonstrate the performance of the proposed algorithm under different simulated conditions, which use the experimentally identified NLOS models. Then we apply the sensor placement algorithm to a real-world cluttered environment for both 2D and 3D settings. Optimized anchor configurations are obtained in simulation and we demonstrate the effectiveness of the designed placement through the improvement of real-world localization performance compared to trivial anchor placements. Finally, we discuss the difference between the theoretical analysis and the real-world localization performance. 

\subsection{Simulation Results}
\begin{figure}[tb]
  \begin{center}
    \includegraphics[width=0.49\textwidth]{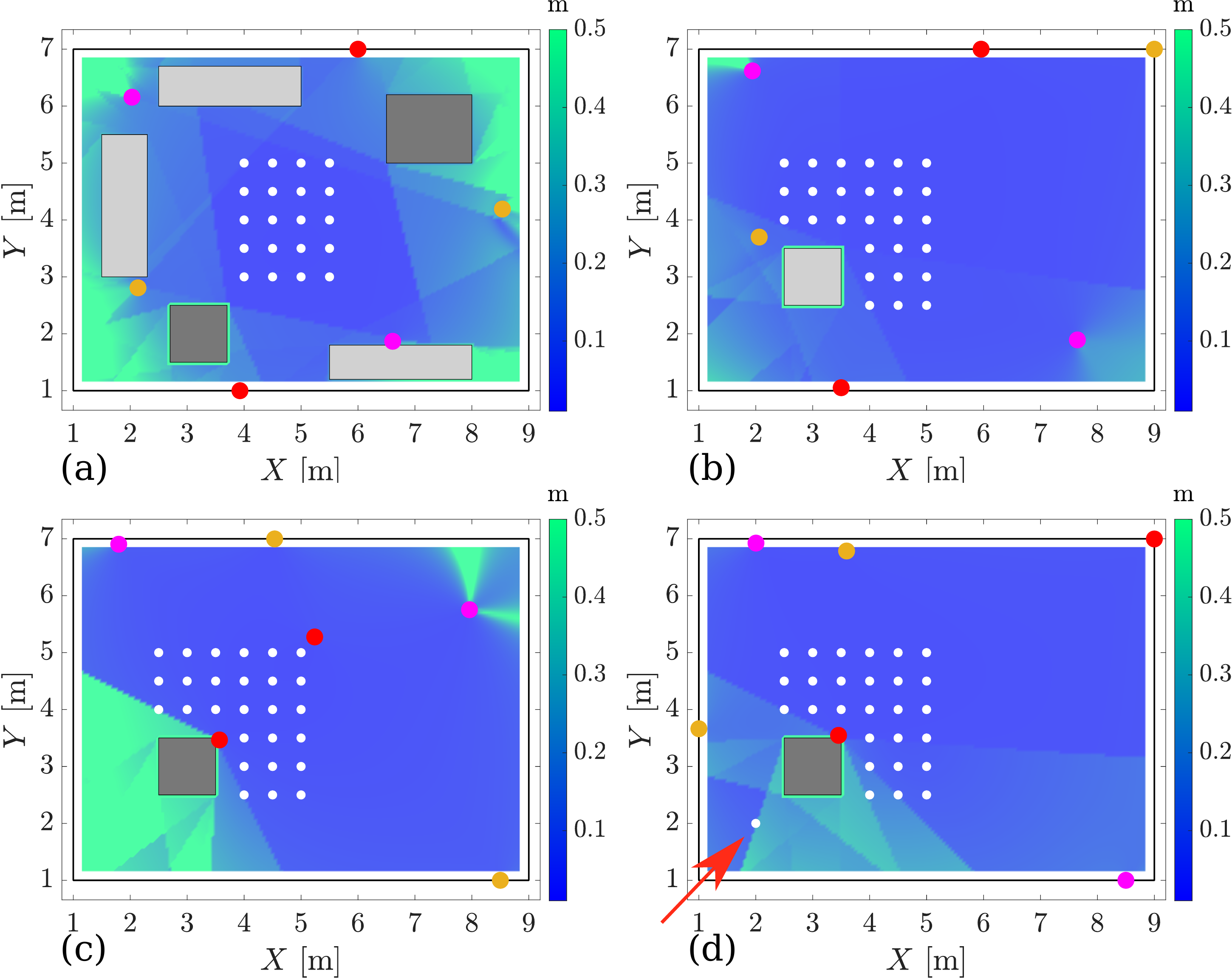}
  \end{center}
  \caption{Simulation results for 2D scenarios. The anchors and sample points are indicated by colored and white dots, respectively, with anchor pairs indicated by the same color. Non-metal and metal obstacles are represented by light and dark gray boxes, respectively. The optimized anchor placements balance the effects of anchor geometry and NLOS measurement biases to improve localization performance. 
  }
  \label{fig:sim_res}
\end{figure}

We present the simulation results in 2D through heatmaps where lower RMSE is indicated with darker color. We first apply the BCM-based anchor placement algorithm (Algorithm~1) to a simple 2D setup, where the region of interest $\Phi$, represented by solid white dots, lies within common and metal obstacles. We allowed anchors to be placed anywhere in the non-obstacle area. 
Good localization performance can be achieved in this case and the optimized anchor configuration, shown in Figure~\ref{fig:sim_res}a, is not trivial. 

We then demonstrate the performance of Algorithm~1 in more challenging conditions with sample points close to obstacles. As shown in the Figure~\ref{fig:sim_res}b, when the obstacle is non-metal, the algorithm balances the effects of NLOS errors and the anchor geometry. When the obstacle is metal (see Figure~\ref{fig:sim_res}c), considering the large NLOS errors induced by such obstacles, the algorithm avoids all NLOS conditions. In Figure~\ref{fig:sim_res}d, we add one extra sample point at $(2,2)$ behind the metal obstacle (highlighted with a red arrow). In this scenario, no anchor configuration can cover all the sample points while avoiding NLOS. The proposed algorithm sacrifices the localization accuracy at the sample points on the right so as to cover the extra sample point. 

\begin{figure}[!b]
  \begin{center}
    \includegraphics[width=0.4\textwidth]{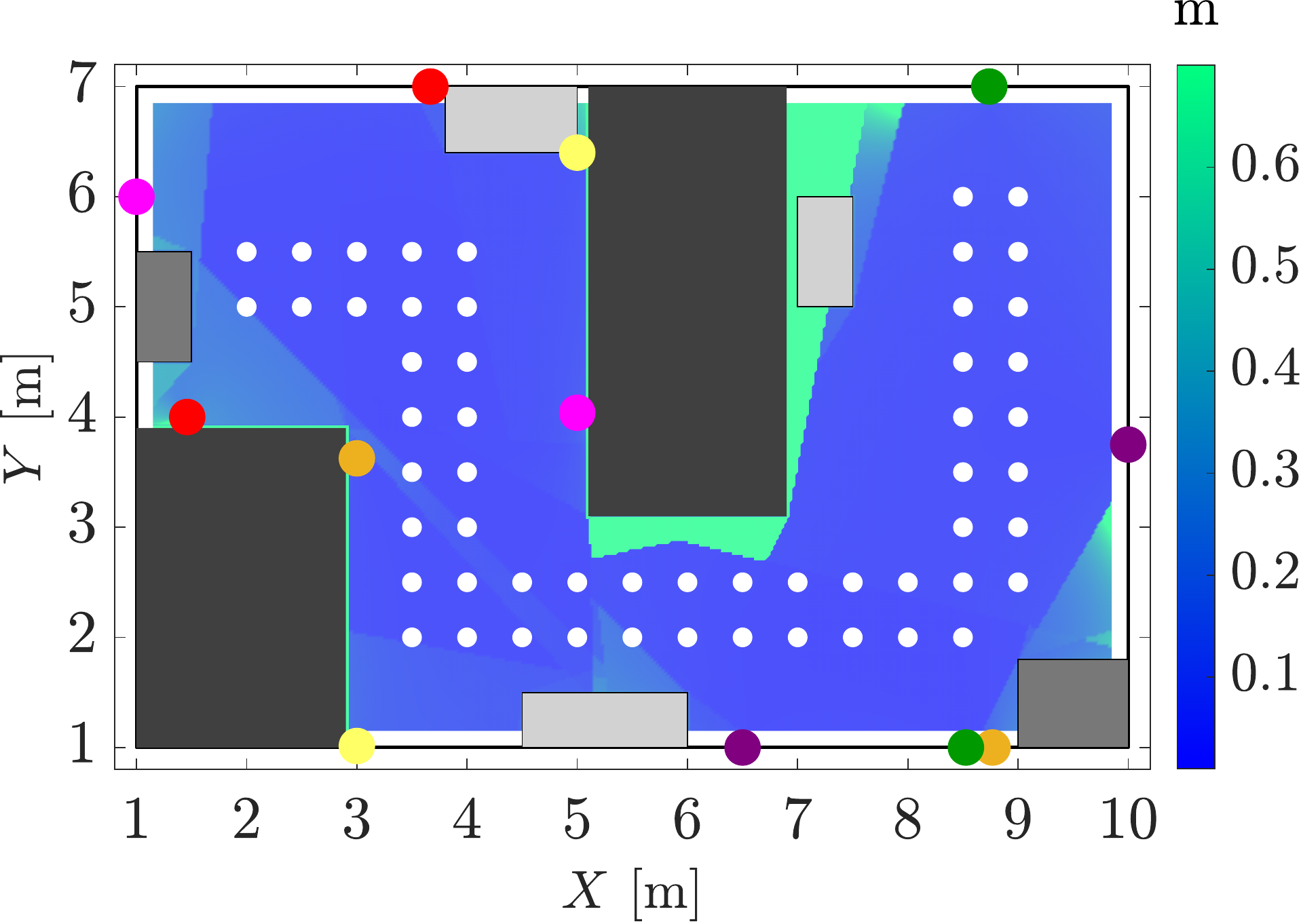}
  \end{center}
  \caption{UWB TDOA-based localization system design results. Solid walls are indicated as black boxes. Twelve anchors are required to meet an accuracy of $0.05$ meter average RMSE over the sample points (white dots).}
  \label{fig:system-design}
\end{figure}

Next, we demonstrate the performance of the proposed TDOA localization system design algorithm. We consider a robot navigation task in a complicated indoor environment with solid walls and various obstacles as shown in Figure~\ref{fig:system-design}. The goal is to design an UWB TDOA-based localization system to provide an accuracy of $0.05$ meter average RMSE over the pre-defined region (indicated by sample points), where the anchors are constrained to be placed on the boundary of the space and walls.
In this complex system design problem, an intuitive anchor placement is generally hard to come up with, especially for TDOA-based hyperbolic localization. Consequently, it is necessary to design the anchor placement through Algorithm~2.
As demonstrated in Figure~\ref{fig:system-design}, $12$ anchors are required to meet the accuracy requirement, leading to around $0.04$ meter average RMSE over the sample points.

Finally, we evaluate the optimization performance for different random initial anchor positions (20 initial conditions for each scenario in Figure~\ref{fig:sim_res}, 10 for the system design scenario in Figure~\ref{fig:system-design}). 
The simulation results show that the variation in the position accuracy due to the random initialization is a magnitude (millimeters) lower than the average position error (centimeter).  Additionally, the number of required anchor pairs computed by Algorithm~2 is the same among all the simulation runs. Detailed simulation results are provided in the supplementary material\hyperlink{fn:supplementary}{$^1$}.

\begin{figure}[tb]
  \begin{center}
    \includegraphics[width=0.41\textwidth]{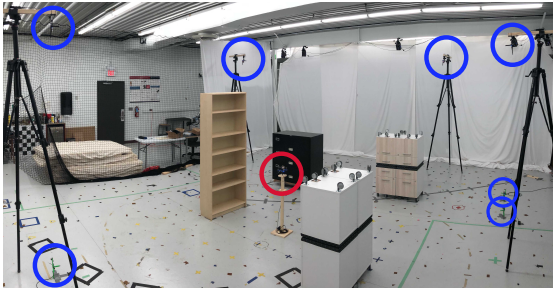}
  \end{center}
  \caption{Optimized 3D anchor placement in a cluttered environment with one metal cabinet (black) and three non-metal obstacles. The tag and anchors are indicated by red and blue circles, respectively.}
  \label{fig:exp-photo}
\end{figure}

\subsection{Experimental Results}
To validate the sensor placement design in realistic settings, we conducted real-world experiments in both 2D and 3D. One metal cabinet and three non-metal obstacles were placed in a $6m\times7m\times2.65m$ indoor space as shown in Figure~\ref{fig:exp-photo}. We used DW1000 UWB modules from Bitcraze's LPS to set up the UWB TDOA-based localization systems. The UWB modules are configured in decentralized mode (TDOA3) and the ground truth positions are provided by a motion capture system comprising of ten Vicon cameras. Since the main focus of this work is the localization performance of a given sensor placement design, we use multilateration~\cite{fang1986trilater} as the position estimator in our experimental validation. In both 2D and 3D experiments, we placed an UWB tag at each sample point to collect UWB measurements. The low-cost DW1000 UWB radio is known to have LOS measurement biases related to the antenna radiation pattern~\cite{zhao2021learning}. Therefore, we first map the LOS biases at each sample point with no obstacles in the space. Then, for the UWB measurements collected in the presence of obstacles, we reject large multi-path outliers with measurement errors larger than one meter and compensate for the LOS biases. In practice, the LOS biases can be identified in a lab setting and are shown to be generalizable to other anchor settings~\cite{zhao2021learning}. 
Moreover, the effects of LOS biases for more advanced UWB systems are usually negligible. At each sample point, the tag position is computed through multilateration.

\begin{figure}[b]
  \begin{center}
   \begin{tikzpicture}
    \centering
    \node[inner sep=0pt, opacity=1] (anchor) at (0.0,0.0){\includegraphics[width=.48\textwidth]{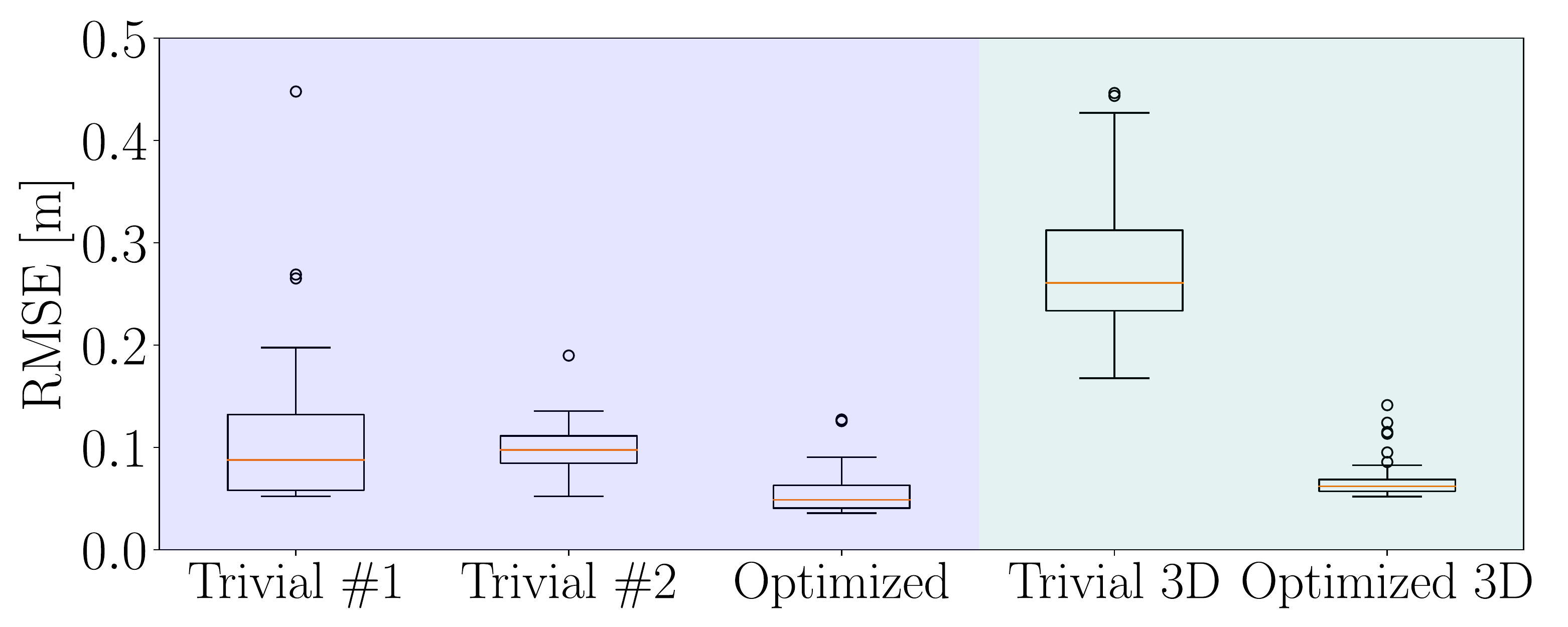}};
    \node[text width=2.5cm, black] at (-0.4, 1.2) {\small 2D Exp.};
    \node[text width=2.5cm, black] at (3.4, 1.2) {\small 3D Exp.};
    \end{tikzpicture}
  \end{center}
 \caption{Root-mean-square error (RMSE) of the multilateration results in both 2D and 3D experiments. The optimized anchor placements achieve around $46.77\%$ and $75.67\%$ RMSE reduction in 2D and 3D, respectively, compared to the trivial placements. }
  \label{fig:boxplot}
\end{figure}
In 2D experiments, we select a challenging region of interest between the obstacles (indicated by the red points in Figure~\ref{fig:experimental_results}a-c) for anchor placement design. Conventional anchor placement design either places anchors in the corner or uniformly distributes the anchors on the boundary of the space, which we refer to as trivial placement $\#1$ and $\#2$, respectively. We apply Algorithm~1 with trivial placement $\#1$ as the initial guess. The corresponding multilateration results for the trivial and optimized anchor configurations are shown in Figure~\ref{fig:experimental_results}a-c. It can be observed that the localization performance in the two trivial placements deteriorates greatly due to the presence of obstacles. With the optimized anchor configuration, we are able to achieve a localization accuracy of $0.058$ meters, which represents an average RMSE reduction of $51.01\%$ and $42.53\%$ compared to the trivial placements $\#1$ and $\#2$, respectively. 

For 3D experiments, we compare the localization performance of the anchor positions computed from Algorithm~1 and a trivial placement. We sample $32$ points at $1m$ and $0.7m$ height. The anchor placement is constrained to the boundary of the 3D space. The multilateration results of the trivial and optimized 3D anchor configurations are shown in Figure~\ref{fig:experimental_results}d-e, respectively. For the trivial 3D anchor placement, the multilateration results show large errors over the sample points. Using the trivial placement as initial guess for optimization, we achieved a $75.67\%$ improvement in the average RMSE (from $0.291m$ to $0.071m$) with the optimized anchor placement. The multilateration RMSE errors in both 2D and 3D experiments are summarized in Figure~\ref{fig:boxplot}. 
\begin{figure}[t]
  \begin{center}
    \begin{tikzpicture}
    \centering
    \node[inner sep=0pt, opacity=1] (anchor) at (0.0,0.0){\includegraphics[width=.48\textwidth]{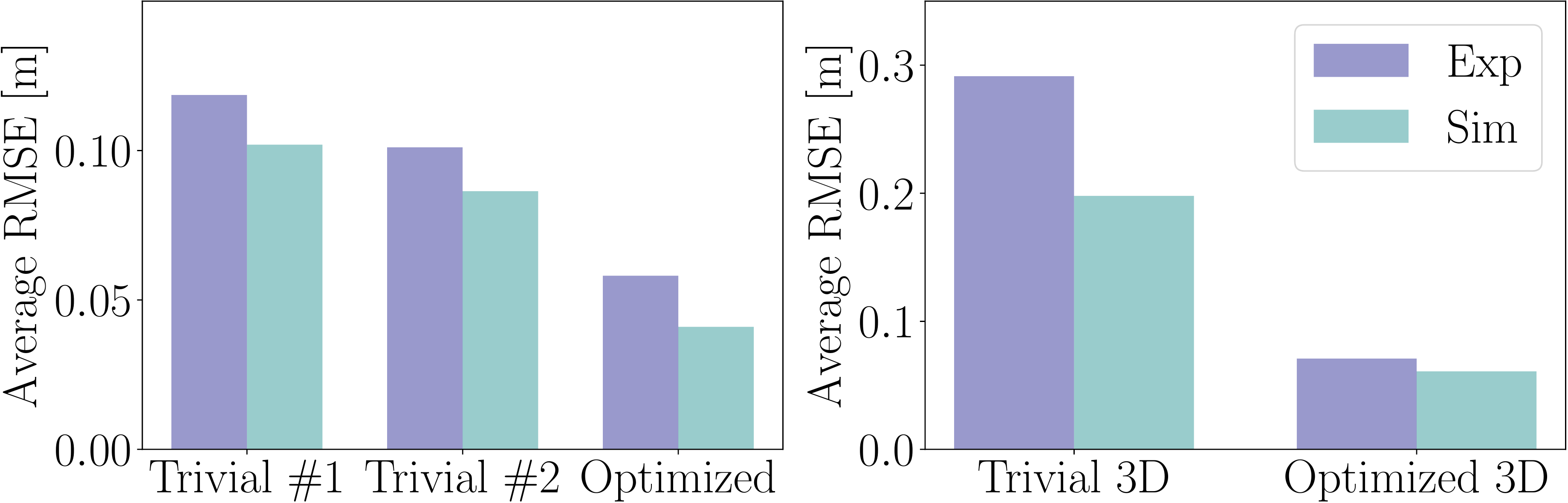}};
    \end{tikzpicture}
  \end{center}
  \caption{Simulation and experimental results in 2D (on the left) and 3D (on the right). The RMSE results in simulation and reality show similar trends.}
  \label{fig:barplot}
\end{figure}

\begin{figure*}[t]
  \begin{center}
    \begin{tikzpicture}
    \centering
    \node[inner sep=0pt, opacity=1] (anchor) at (0.0,0.0){\includegraphics[width=.85\textwidth]{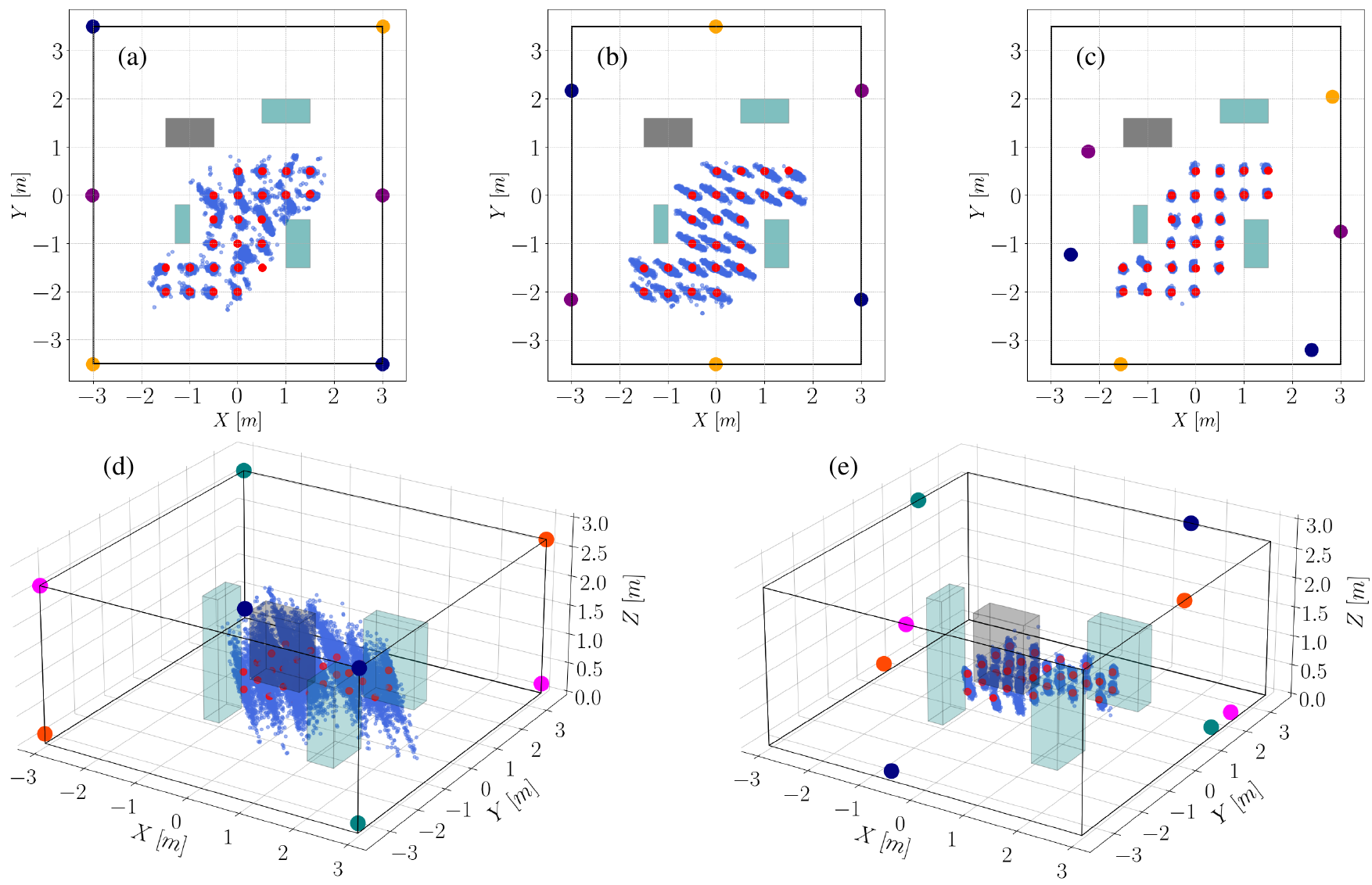}};
    \end{tikzpicture}
  \end{center}
\caption{Multilateration results (light blue dots) in real-world experiments. The non-metal and metal obstacles are indicated as gray and teal boxes. The ground truth sample points are indicated as red points. The trivial anchor placements are selected in the corner (2D trivial placement $\#1$,~(a) and 3D trivial placement~(d)) and uniformly distributed on the boundary (2D trivial placement $\#2$,~(b)) of the space. The optimized anchor placements in 2D~(c) and 3D~(e) balance the effects of anchor geometry and NLOS measurement biases, leading to improved localization performance over selected sample points.}
\label{fig:experimental_results}
\end{figure*}

Finally, we compare the average RMSE values from simulation and real-world experiments for 2D and 3D anchor configurations in Figure~\ref{fig:barplot}. It can be observed that the simulation and experimental results show similar trends, verifying that our NLOS models capture the real-world effects reasonably well. In reality, there exist small multi-path measurements that cannot be easily distinguished and rejected, which degrade localization performance. Moreover, the imperfect models for the UWB NLOS measurements also cause gaps between simulation and reality. We observe that the optimized 3D anchor placement shows the smallest RMSE difference between simulation and experimental results. The reason is that the optimized 3D anchor placement nearly avoids all NLOS conditions and thus the model imperfection does not affect the simulation results.

\section{Conclusions}
In this work, we developed a sensor placement algorithm to optimize the UWB anchor configuration for a decentralized TDOA-based localization system w.r.t. a region of interest in cluttered environments. We use the MSE metric to evaluate the performance of a TDOA localization system, which enables us to balance the effects of anchor geometry and the NLOS measurement biases. For sensor placement optimization, we constructed comprehensive and efficient NLOS error models and proposed a BCM-based optimization algorithm to optimize the anchor placement. We further used the anchor placement optimization for localization system design to compute a minimal number of UWB  anchors and their corresponding positions to meet a pre-defined accuracy requirement. We conducted real-world experiments in both 2D and 3D in a cluttered environment and demonstrated that the localization performance improves by around $46.77\%$ in 2D and $75.67\%$ in 3D with the proposed optimized placement of anchors compared to trivial placements. 





\newpage
\onecolumn
\section{\large{Supplemental Material}}

This report provides detailed information about UWB TDOA models, Fisher Information Matrix (FIM) derivation, and the proposed system design algorithm to support the paper ``\textit{Finding the Right Placement: Sensor Placement for Time Difference of Arrival Localization in Cluttered Indoor Environments}".

\subsection{UWB TDOA Measurement Model}
In this subsection, we provide detailed information about the UWB TDOA measurement models under various NLOS conditions and the FIM derivation.
\subsubsection{General UWB TDOA Measurement Model}
\label{sec:general_mm}
The TDOA measurements are modelled as 
\begin{equation}
\label{eq:noisy-tdoa-model}
   d_{ij}(\mathbf{p},\bm{a}_i,\bm{a}_j)  = \bar{d}_{ij}(\mathbf{p},\bm{a}_i,\bm{a}_j) +
   \varepsilon_{ij} + b_{ij,\textrm{nlos}}, 
\end{equation}
where $\varepsilon_{ij} \sim f_{\mathcal{N}}$ is the zero-mean Gaussian measurement noise with variance $\sigma^2$, $\mathbf{p}$ is the point of interest, and $\mathbf{a}_i$ and $\mathbf{a}_j$ are the anchor positions. We obtain the variance $\sigma^2$ from the LOS experiment.

The UWB NLOS measurement bias is modeled as follows:
\begin{equation}
\label{eq:general_tdoa_mm}
b_{ij,\textrm{nlos}}  = Y_i b_i + Y_j b_j +  Y_{ij} b_{ij}, (i,j) \in \Gamma,
\end{equation}
where $\{b_{i}, b_{j}, b_{ij}\}$ are random variables that represent the measurement errors induced by NLOS events $\{\bar{L}_i, \bar{L}_j, \bar{L}_{ij}\}$, respectively. The independent random variables $Y_i$, $Y_j$, and $Y_{ij}$ follow Bernoulli distributions, which take the value $1$ with the probability of the corresponding NLOS events. Given positions of obstacles, anchors $\bm{a}$ and the tag $\mathbf{p}$, NLOS conditions can be detected through ray tracing. Therefore, $Y_i$, $Y_j$, and $Y_{ij}$ take the value $1$ when the corresponding NLOS event is detected and $0$ otherwise. 

For clarity, we drop the common and severe NLOS category superscripts while modeling. As presented in the paper, for the NLOS event $(L_i,\bar{L}_j,L_{ij})$ we model the NLOS bias as log-normal distributions and the probability density of the measurement errors can be written as
\begin{equation}
    f(\Delta d_{ij} | L_i, \bar{L}_j, L_{ij}) = (f_{ln\mathcal{N},j}*f_{\mathcal{N}})(\Delta d_{ij}),
\end{equation}
where $*$ operator indicates convolution of probability distributions. Analogously, we can write the probability density of measurement errors for event $(\bar{L}_i, L_j, L_{ij})$ as
\begin{equation}
    f(\Delta d_{ij} | \bar{L}_i, L_j, L_{ij}) = (f^{-}_{ln\mathcal{N},i}*f_{\mathcal{N}})(\Delta d_{ij}),
\end{equation}
where we denote the negative log-normal distribution as $f^{-}_{ln\mathcal{N},i}(\tau) = f_{ln\mathcal{N},i}(-\tau)$ with $\tau \in\mathbb{R}$.

For the NLOS event $(L_i,L_j,\bar{L}_{ij})$, we observe that the measurement errors remain zero-mean for both common and severe NLOS (see Figure~1b-c in the paper). Hence, we indicate the probability density of $b_{ij}$ as $f_{ij}$ and model the measurement errors under $(L_i, L_j, \bar{L}_{ij})$ as
\begin{equation}
    f(\Delta d_{ij} | L_i, L_j, \bar{L}_{ij}) = (f_{ij}*f_{\mathcal{N}})(\Delta d_{ij}).
\end{equation}
With $f_{ln\mathcal{N},j}$, $f^{-}_{ln\mathcal{N},i}$, and $f_{ij}$, we can derive the probability density function $f_{b_{ij},\textrm{nlos}}$ for $b_{ij,\textrm{nlos}}$ in Equation~\eqref{eq:general_tdoa_mm} for different NLOS conditions through convolution, and obtain an accurate measurement error model
\begin{equation}
    f(\Delta d_{ij}) = (f_{b_{ij},\textrm{nlos}} * f_{\mathcal{N}})(\Delta d_{ij}).
\end{equation}

The probability density functions of measurement errors for the eight NLOS events can be summarized as follows:
\begin{equation}
\label{eq:eight_cond_models}
    \begin{split}
        f(\Delta d_{ij}|L_i,L_j,L_{ij})       &= f_{\mathcal{N}}(\Delta d_{ij}) \\
        f(\Delta d_{ij}|L_i,\bar{L}_j,L_{ij}) &= (f_{\mathcal{N}} * f_{ln\mathcal{N},j})(\Delta d_{ij})  \\
        f(\Delta d_{ij}|\bar{L}_i,L_j,L_{ij}) &= (f_{\mathcal{N}} * f^{-}_{ln\mathcal{N},i})(\Delta d_{ij})  \\
        f(\Delta d_{ij}|\bar{L}_i,\bar{L}_j,L_{ij}) &= (f_{\mathcal{N}} * f_{ln\mathcal{N},j} * f^{-}_{ln\mathcal{N},i})(\Delta d_{ij}) \\
        f(\Delta d_{ij}|L_i,L_j,\bar{L}_{ij}) &= (f_{\mathcal{N}} * f_{ij})(\Delta d_{ij})  \\
        f(\Delta d_{ij}|L_i,\bar{L}_j,\bar{L}_{ij}) &= (f_{\mathcal{N}} * f_{ij} * f_{ln\mathcal{N},j})(\Delta d_{ij})  \\
        f(\Delta d_{ij}|\bar{L}_i,L_j,\bar{L}_{ij}) &= (f_{\mathcal{N}} * f_{ij} * f^{-}_{ln\mathcal{N},i})(\Delta d_{ij})  \\
        f(\Delta d_{ij}|\bar{L}_i,\bar{L}_j,\bar{L}_{ij}) &= (f_{\mathcal{N}} * f_{ij} * f_{ln\mathcal{N},j} * f^{-}_{ln\mathcal{N},i})(\Delta d_{ij}).
    \end{split}
\end{equation}

Note, the errors induced by common and severe NLOS conditions, $f_{ln\mathcal{N},j}, f^{-}_{ln\mathcal{N},i}$, and $f_{\mathcal{N},ij}$ in Equation \eqref{eq:eight_cond_models}, are modeled separately to achieve representative error models for various indoor NLOS scenarios.

\subsubsection{Efficient TDOA Measurement Model}
\label{sec:efficient_mm}
To reduce the computational burden of anchor placement optimization, we approximate the NLOS measurement models as closed-form Gaussian distributions, which have analytical forms in Fisher Information Matrix (FIM) calculation.
\begin{figure}[tb]
  \begin{center}
    \includegraphics[width=0.99\textwidth]{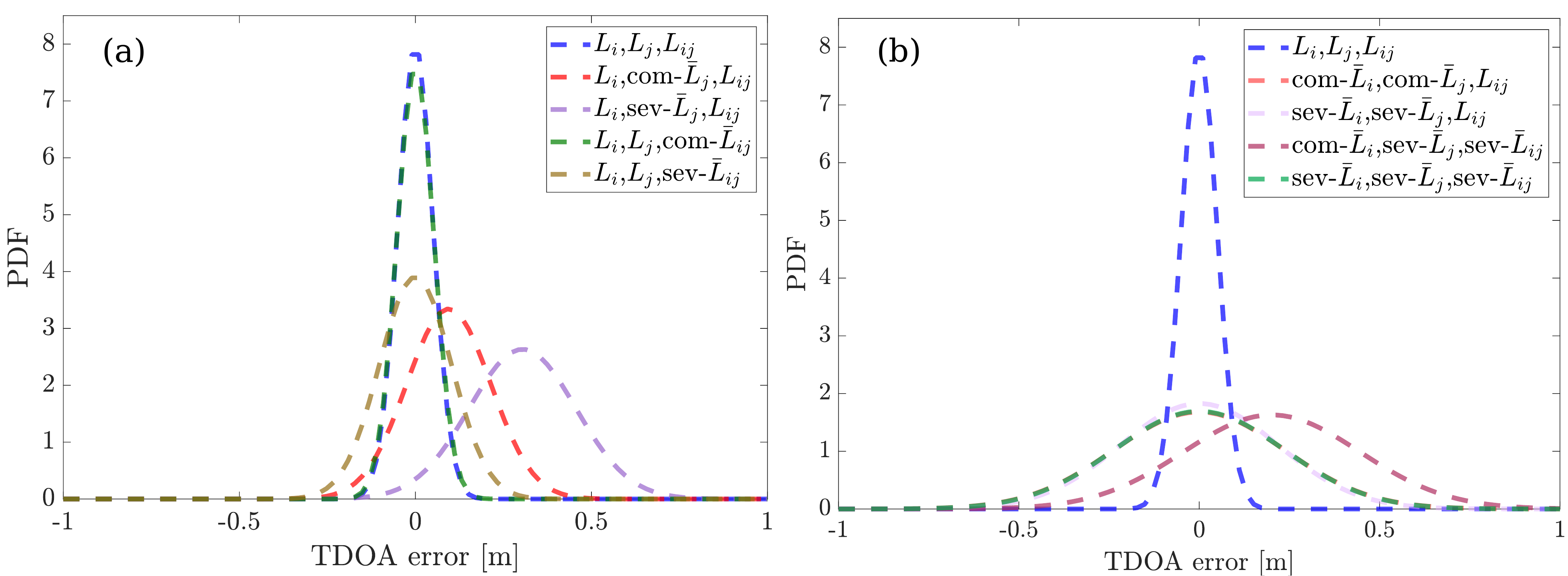}
  \end{center}
  \caption{A subset (8 of the total 27) of proposed UWB NLOS error models. The probability density function for the LOS measurement error (indicated aby the blue dashed line) is shown for comparison. }
  \label{fig:nlos-model}
\end{figure}

For the NLOS events $(L_i,\bar{L}_j,L_{ij})$, $(\bar{L}_i,L_j,L_{ij})$, and $(\bar{L}_i, \bar{L}_j,L_{ij})$, we approximate the probability density function of the NLOS bias by finding the closest Gaussian distributions, which we obtain through minimizing the Kullback-Leibler (KL) divergence. As numerically validated in~\cite{prorok2012online}, the Gaussian approximation $f_{\tilde{\mathcal{N}},j}$ is a good match for the convolution $f_{ln\mathcal{N},j} * f^{-}_{ln\mathcal{N},i}$  corresponding to the NLOS event $(\bar{L}_i, \bar{L}_j,L_{ij})$. For the long-tail distribution in the severe NLOS event $(L_i, L_j,\bar{L}_{ij})$, we construct a Gaussian distribution $f_{\mathcal{N},ij}$ with the mean and variance computed from experimental data to represent the uncertainty of the NLOS biases. Therefore, the eight NLOS events are approximated as follows:

\begin{equation}
\label{eq:eight_cond_approx}
    \begin{split}
        f(\Delta d_{ij}|L_i,L_j,L_{ij})       &= f_{\mathcal{N}}(\Delta d_{ij}), \\  
        f(\Delta d_{ij}|L_i,\bar{L}_j,L_{ij}) &= (f_{\mathcal{N}} * f_{ln\mathcal{N},j})(\Delta d_{ij}) \approx (f_{\mathcal{N}} * f_{\tilde{\mathcal{N}},j})(\Delta d_{ij}), \\        
        f(\Delta d_{ij}|\bar{L}_i,L_j,L_{ij}) &= (f_{\mathcal{N}} * f^{-}_{ln\mathcal{N},i})(\Delta d_{ij}) \approx (f_{\mathcal{N}} * f^{-}_{\tilde{\mathcal{N}},i})(\Delta d_{ij}),  \\  
        f(\Delta d_{ij}|\bar{L}_i,\bar{L}_j,L_{ij}) &= (f_{\mathcal{N}} * f_{ln\mathcal{N},j} * f^{-}_{ln\mathcal{N},i})(\Delta d_{ij}) \approx (f_{\mathcal{N}} * f_{\tilde{\mathcal{N}}})(\Delta d_{ij}), \\  
        f(\Delta d_{ij}|L_i,L_j,\bar{L}_{ij}) &= (f_{\mathcal{N}} * f_{ij})(\Delta d_{ij}) \approx  (f_{\mathcal{N}} * f_{\mathcal{N},ij})(\Delta d_{ij}),\\     
        f(\Delta d_{ij}|L_i,\bar{L}_j,\bar{L}_{ij}) &= (f_{\mathcal{N}} * f_{ij} * f_{ln\mathcal{N},j})(\Delta d_{ij}) \approx (f_{\mathcal{N}} * f_{\mathcal{N},ij} * f_{\tilde{\mathcal{N}},j})(\Delta d_{ij}), \\ 
        f(\Delta d_{ij}|\bar{L}_i,L_j,\bar{L}_{ij}) &= (f_{\mathcal{N}} * f_{ij} * f^{-}_{ln\mathcal{N},i})(\Delta d_{ij}) \approx (f_{\mathcal{N}} * f_{\mathcal{N},ij} * f^{-}_{\tilde{\mathcal{N}},i})(\Delta d_{ij}), \\  
        f(\Delta d_{ij}|\bar{L}_i,\bar{L}_j,\bar{L}_{ij}) &= (f_{\mathcal{N}} * f_{ij} * f_{ln\mathcal{N},j} * f^{-}_{ln\mathcal{N},i})(\Delta d_{ij}) \approx (f_{\mathcal{N}} * f_{\mathcal{N},ij} * f_{\tilde{\mathcal{N}} })(\Delta d_{ij}).  
    \end{split}
\end{equation}
Considering three possible conditions (LOS, common NLOS, and severe NLOS) between either two UWB radios, we have in total $27$ TDOA models for anchor placement analysis. We demonstrated eight representative NLOS error models in Figure~\ref{fig:nlos-model}. It can be observed that the probability density function of common NLOS errors in event ($L_i$,$L_j$,$\bar{L}_{ij}$) (indicated as green dashed line ($L_i$,$L_j$,com-$\bar{L}_{ij}$) in Figure~\ref{fig:nlos-model}a) is similar to $f(\Delta d|L_i,L_j,L_{ij})$ since the common NLOS event $\bar{L}_{ij}$ has little impact on the TDOA measurements. When the two anchor-tag pairs are affected by the same NLOS conditions, we get high variance distributions which dominates the NLOS condition between the anchor-anchor pair. 

\subsection{Fisher Information Matrix Derivation}
In this subsection, we provide a detailed derivation of the Fisher Information Matrix (FIM) when the TDOA measurements follow a Gaussian distribution. 

We define the anchor positions $\bm{a}=\{\bm{a}_i,\bm{a}_j\}$ and the tag position $\mathbf{p}$ in 2D as follows:
\begin{equation}
        \bm{a}_i = [a_{1i}, a_{2i}]^T, ~~~
        \bm{a}_j = [a_{1j}, a_{2j}]^T, ~~~
        \mathbf{p} = [p_1, p_2]^T.
\end{equation}

The TDOA measurement $d_{ij}$ is assumed to follow a Gaussian distribution with the probability density function
\begin{equation}
    f(d_{ij};\bm{a},\mathbf{p}) = \frac{1}{\sigma \sqrt{2\pi}} \exp\left(-\frac{1}{2\sigma^2}\left(d_{ij}-\Big(\|\mathbf{p}-\bm{a}_j\| - \|\mathbf{p}-\bm{a}_i\| + b_{ij,\textrm{nlos}}\Big)\right)^2 \right),
\end{equation}
where $\sigma$ indicates the standard derivation, $b_{ij,\textrm{nlos}}$ denotes the NLOS measurement error, and $ \|\cdot\|$ indicates the $\ell_2$ norm. For clarity, we define the ideal TDOA measurement as
\begin{equation}
\bar{d}_{ij}(\bm{a},\mathbf{p}) = \|\mathbf{p}-\bm{a}_j\| - \|\mathbf{p}-\bm{a}_i\|. 
\end{equation}

The derivation for the $(1,2)$ element of the FIM $\mathcal{I}(\mathbf{p})$ is: 
\begin{equation}
\begin{split}
    &\left[\mathcal{I}(\mathbf{p})\right]_{1,2} \\
    =& - \mathbb{E}\left[\frac{\partial^2 \log f(d_{ij};\bm{a},\mathbf{p})}{\partial p_1 \partial p_2}\right] \\
    =& -\mathbb{E}\bigg\{\frac{\partial^2}{\partial p_1, \partial p_2}\left[-\frac{1}{2\sigma^2}\Big(d_{ij} - b_{ij,\textrm{nlos}} - \bar{d}_{ij}(\bm{a},\mathbf{p}) \Big)^2\right]\bigg\}   \\
    =& \frac{1}{\sigma^2} \mathbb{E} \bigg\{\frac{\partial^2}{\partial p_1, \partial p_2} \Big[\frac{1}{2} \left((d_{ij} - b_{ij,\textrm{nlos}})^2 - 2(d_{ij} - b_{ij,\textrm{nlos}})\bar{d}_{ij}(\bm{a},\mathbf{p}) +  \bar{d}_{ij}(\bm{a},\mathbf{p})^2\right)\Big] \bigg\}  \\
    =&  \frac{1}{\sigma^2} \mathbb{E} \bigg\{\frac{\partial}{\partial p_{2}} \Big[ -(d_{ij}-b_{ij,\textrm{nlos}})\left(\frac{p_1 - a_{1j}}{\|\mathbf{p}-\bm{a}_j\|} - \frac{p_1 - a_{1i}}{\|\mathbf{p}-\bm{a}_i\|}\right) + \bar{d}_{ij}(\bm{a},\mathbf{p})\left(\frac{p_1 - a_{1j}}{\|\mathbf{p}-\bm{a}_j\|} - \frac{p_1 - a_{1i}}{\|\mathbf{p}-\bm{a}_i\|}\right) \Big]  \bigg\} \\
    =& \frac{1}{\sigma^2} \mathbb{E} \bigg\{ (d_{ij}-b_{ij,\textrm{nlos}}) \Big(\frac{(p_1-a_{1j})(p_2-a_{2j})}{(\|\mathbf{p}-\bm{a}_j\|)^3}-\frac{(p_1-a_{1i})(p_2-a_{2i})}{(\|\mathbf{p}-\bm{a}_i\|)^3}\Big) \bigg.\\  
    & ~~~~~ + \left(\frac{(p_2-a_{2j})}{\|\mathbf{p}-\bm{a}_j\|} - \frac{(p_2-a_{2i})}{\|\mathbf{p}-\bm{a}_i\|}\right)\left(\frac{(p_1-a_{1j})}{\|\mathbf{p}-\bm{a}_j\|} - \frac{(p_1-a_{1i})}{\|\mathbf{p}-\bm{a}_i\|}\right)\\
    & ~~~~~ - \left(\|\mathbf{p}-\bm{a}_j\| - \|\mathbf{p}-\bm{a}_i\|\right) \Big(\frac{(p_1-a_{1j})(p_2-a_{2j})}{(\|\mathbf{p}-\bm{a}_j\|)^3}-\frac{(p_1-a_{1i})(p_2-a_{2i})}{(\|\mathbf{p}-\bm{a}_i\|)^3}\Big) \bigg\}  \\
    =& \frac{1}{\sigma^2} \mathbb{E}\bigg\{d_{ij} - \left(b_{ij,\textrm{nlos}} + (\|\mathbf{p}-\bm{a}_j\|-\|\mathbf{p}-\bm{a}_i\|)\right)\bigg\}\Big(\frac{(p_1-a_{1j})(p_2-a_{2j})}{(\|\mathbf{p}-\bm{a}_j\|)^3}-\frac{(p_1-a_{1i})(p_2-a_{2i})}{(\|\mathbf{p}-\bm{a}_i\|)^3}\Big) \\
    &~~~~~ + \frac{1}{\sigma^2}\left(\frac{(p_2-a_{2j})}{\|\mathbf{p}-\bm{a}_j\|} - \frac{(p_2-a_{2i})}{\|\mathbf{p}-\bm{a}_i\|}\right)\left(\frac{(p_1-a_{1j})}{\|\mathbf{p}-\bm{a}_j\|} - \frac{(p_1-a_{1i})}{\|\mathbf{p}-\bm{a}_i\|}\right)  \\
    =&\frac{1}{\sigma^2}\left(\frac{(p_1-a_{1j})}{\|\mathbf{p}-\bm{a}_j\|} - \frac{(p_1-a_{1i})}{\|\mathbf{p}-\bm{a}_i\|}\right)\left(\frac{(p_2-a_{2j})}{\|\mathbf{p}-\bm{a}_j\|} - \frac{(p_2-a_{2i})}{\|\mathbf{p}-\bm{a}_i\|}\right)
\end{split}
\label{eq:fim_derivation}
\end{equation}

Therefore, the analytical FIM computation in 2D is as follows
\begin{equation}
\renewcommand\arraystretch{2}
\mathcal{I}(\mathbf{p}) = \frac{1}{\sigma^2} 
\begin{bmatrix}
\displaystyle\frac{(p_1-a_{1j})}{\|\mathbf{p}-\bm{a}_j\|} - \displaystyle\frac{(p_1-a_{1i})}{\|\mathbf{p}-\bm{a}_i\|} \\
\displaystyle\frac{(p_2-a_{2j})}{\|\mathbf{p}-\bm{a}_j\|} - \displaystyle\frac{(p_2-a_{2i})}{\|\mathbf{p}-\bm{a}_i\|} 
\end{bmatrix}
\begin{bmatrix}
\displaystyle\frac{(p_1-a_{1j})}{\|\mathbf{p}-\bm{a}_j\|} - \displaystyle\frac{(p_1-a_{1i})}{\|\mathbf{p}-\bm{a}_i\|} &
\displaystyle\frac{(p_2-a_{2j})}{\|\mathbf{p}-\bm{a}_j\|} - \displaystyle\frac{(p_2-a_{2i})}{\|\mathbf{p}-\bm{a}_i\|} 
\end{bmatrix}
\end{equation}
In 3D settings, the analytical form of FIM can be computed in a similar manner.

\subsection{System Design Algorithm}
In this subsection, we provide the system design algorithm based on the optimization problem formulated in the paper. Given the obstacle positions, a pre-defined region of interest $\Phi$, and required localization accuracy $\mathcal{M}_r$, we propose an iterative algorithm based on Algorithm~1 in the paper to find a minimal number of anchors required to meet the localization accuracy $\mathcal{M}_r$ and their corresponding positions. 
\setcounter{algocf}{1}
\begin{algorithm}[b]
  \SetKwInput{KwData}{Input    ~}
  \SetKwInput{KwResult}{Output}
  \KwData{Initial number of anchor pairs $Q_{init}$, maximum number of anchor pairs $Q_{max}$, required localization accuracy (RMSE) $\mathcal{M}_r$ for $\Phi \subset \left(\mathcal{P} \cap \mathcal{O}'\right)$,  initial anchor positions $\Scale[1.0]{\bm{a} \in \mathcal{A}_b}$, and other inputs in Algorithm $1$.}
  \KwResult{Minimal number of anchors needed to achieve the required RMSE $\mathcal{M}_r$ and the anchor positions $\Scale[1.0]{\bm{a}^{\star} \in \mathcal{A}_b}$.}
  $Q = Q_{init}$ \;
  Select $N$ sample points $\mathbf{p}=\{\mathbf{p}_1,\cdots, \mathbf{p}_N\} \in \Phi$ \; 
  Run lines $2-9$ of Algorithm 1\;
  Calculate the $\mathcal{M}(\bm{a}^{\star}_{Q})$ based on $\bm{a}^{\star}_{Q}$ \;
  \While{$\mathcal{M}(\bm{a}^{\star}_{Q}) > \mathcal{M}_r$ or $Q < Q_{max}$}{
      Add one extra pair of anchor,  $Q \gets Q+1$, $\bm{a} \gets \{\bm{a}, \bm{a}_{m+1}, \bm{a}_{m+2}\}, m\gets m+2$\;
      Run lines $2-9$ of Algorithm 1\;
      Calculate the $\mathcal{M}(\bm{a}^{\star}_{Q})$ based on $\bm{a}^{\star}_{Q}$ \;
  }
  \Return A minimal number of anchors required $m^{\star}=2Q^{\star}$ and the corresponding placement $\bm{a}^{\star}$ 
  \caption{UWB TDOA-based localization system design algorithm}
\end{algorithm}
Considering complicated indoor environments with solid walls and other obstacles, we constrain the anchor positions to the boundary of the indoor space. In each iteration, we optimize the anchor placement and add one extra pair of anchors if the desired localization accuracy is not satisfied. The additional anchor pair is initialized randomly on the boundary of the space. The algorithm terminates when the required accuracy is satisfied or no solution can be found with the maximum allowable number of anchors. The proposed system design algorithm is presented in Algorithm~2. 

\subsection{Algorithm Evaluation}
In this subsection, we evaluate the performance of the algorithm for randomized initial anchor positions. We considered the scenarios in the paper (Figure~\ref{fig:sim_res} and~\ref{fig:system-design}) and used randomly generated initial anchor positions. We refer to the scenarios in Figure~\ref{fig:sim_res}a-d and Figure~\ref{fig:system-design} as scenario~$\#$1-4 and multi-room scenario, respectively. 
\begin{figure}[!b]
  \begin{center}
    \includegraphics[width=0.99\textwidth]{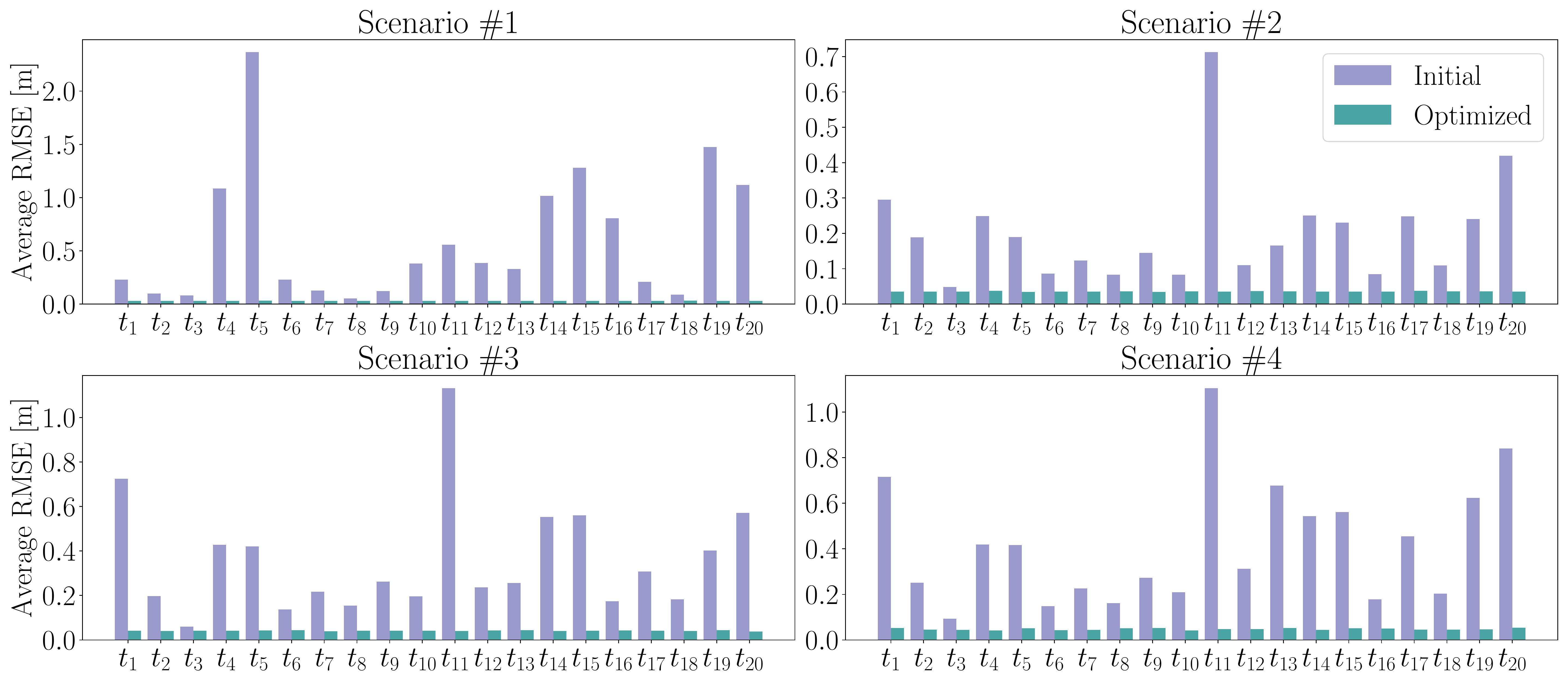}
  \end{center}
  \caption{The average root-mean-square error (RMSE) results with \textit{(i)} randomized initial anchor positions and \textit{(ii)} optimized anchor positions for the four scenarios in the paper (Figure~\ref{fig:sim_res}). For each scenario, we generated $20$ random initial anchor positions and executed the sensor placement optimization with Algorithm 1. It can be observed that the proposed algorithm effectively optimizes the anchor positions to provide good localization accuracy under random initialization.}
  \label{fig:rebuttal_barplot}
\end{figure}
For the general sensor placement problem, there may exist several sensor placements which provide similar localization performance, leading to multiple local optimal solutions. Hence, we assess the optimization performance through the localization accuracy with the optimized anchor positions. 
\begin{figure}[!b]
  \begin{center}
    \includegraphics[width=1.0\textwidth]{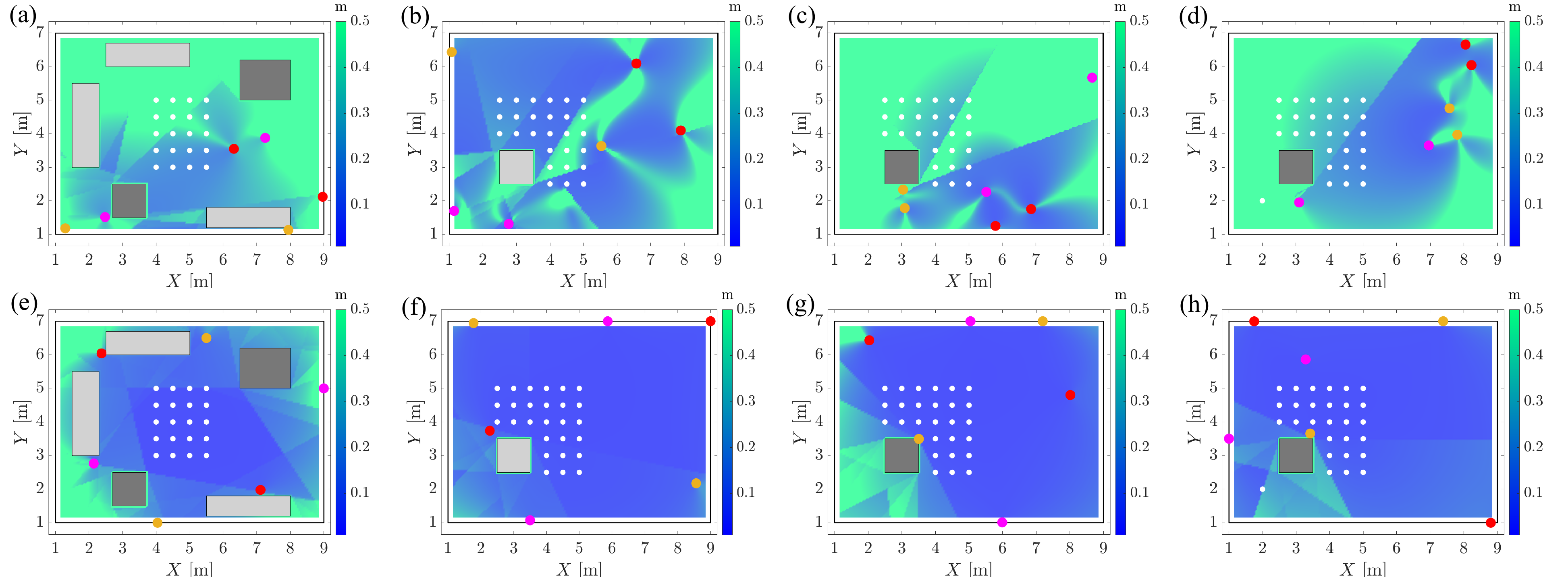}
  \end{center}
  \caption{The figures in the first row (a-d) demonstrate results from one of the $20$ randomly generated initial anchor positions for scenario~$\#$1-4 (Figure~\ref{fig:sim_res}). The figures in the second row (e-h) present the corresponding optimized anchor placements.}
  \label{fig:compare1}
\end{figure}
For scenario~$\#$1-4, we randomly generate $20$ initial anchor positions and use them to perform the proposed sensor placement optimization (Algorithm 1). We summarize the average root-mean-square error (RMSE) of the target points of interest in each simulation test $t_n, n=1,\cdots, 20$ for the \textit{(i)} randomized initial anchor positions and the \textit{(ii)} optimized anchor positions in Figure~\ref{fig:rebuttal_barplot}. It can be observed that the proposed algorithm effectively optimizes the anchor positions to provide good localization accuracy with random initialization. For the multi-room scenario, we tested the localization system design algorithm (Algorithm 2) with $10$ randomised initial conditions. In each case, eight anchors were initialized randomly on the boundary of the space. In all $10$ optimization runs, the proposed algorithm required $12$ anchors to meet the $0.05$~m accuracy requirement. We show the initial and the corresponding optimized placement of the anchors for scenario~$\#$1-4 from one such experiment in Figure~\ref{fig:compare1}. Similarly, one example for multi-room is presented in Figure~\ref{fig:compare2}.
\begin{figure}[t!]
  \begin{center}
    \includegraphics[width=0.9\textwidth]{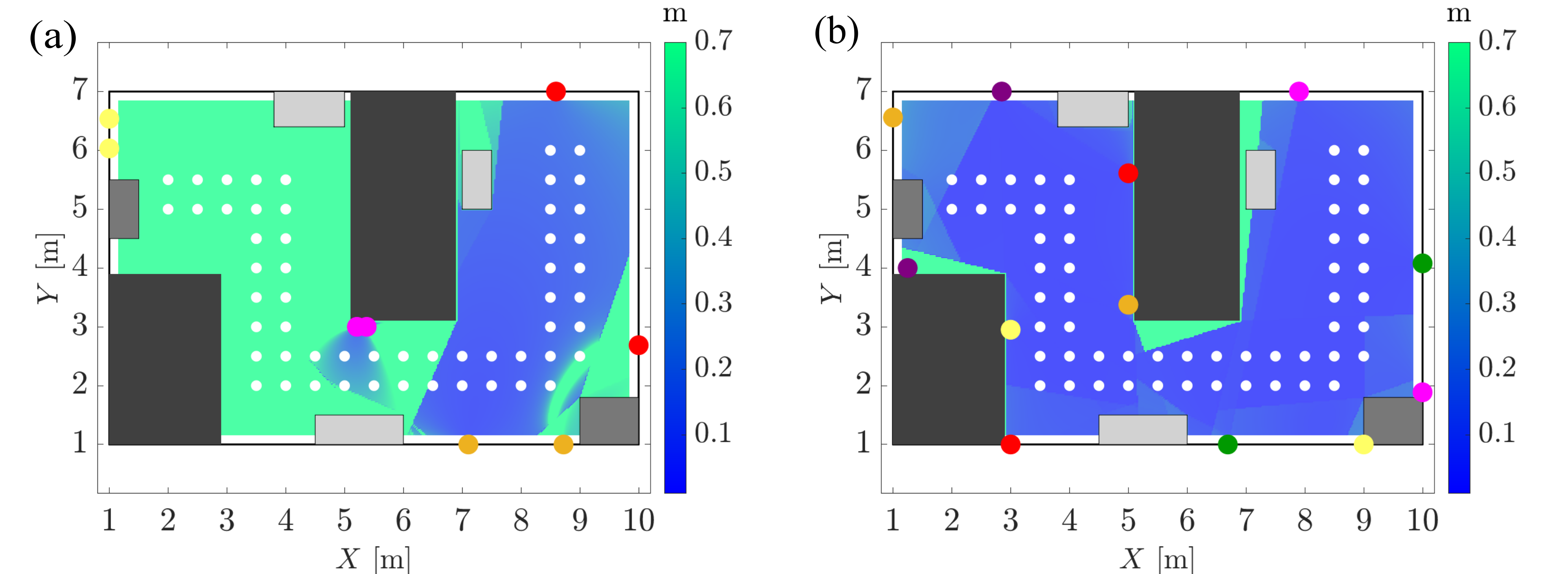}
  \end{center}
  \caption{Comparison of the localization results with (a) the randomly generated anchor placement and (b) the optimized anchor placement in the multi-room scenario. Eight anchors are randomly placed on the boundary of the space as the initial condition. The proposed algorithm successfully optimizes the localization performance over the sample points.   }
  \label{fig:compare2}
\end{figure}

\begin{figure}[t!]
  \begin{center}
    \includegraphics[width=0.7\textwidth]{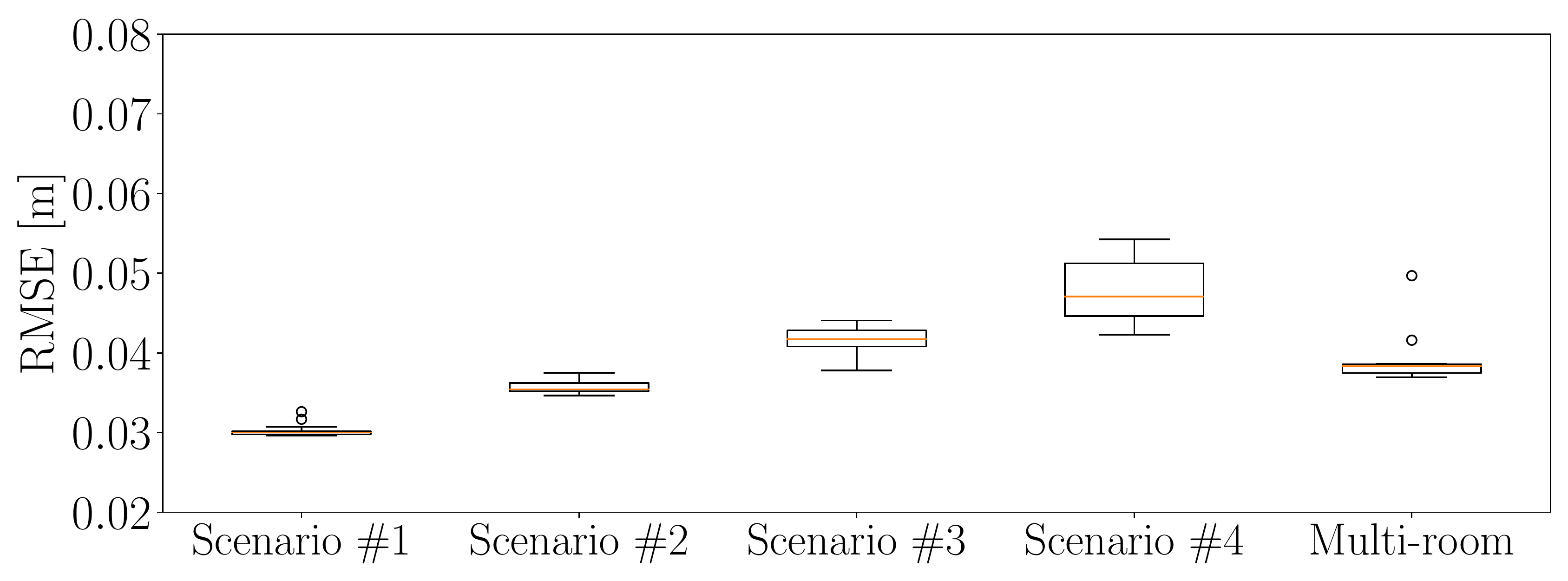}
  \end{center}
  \caption{The average root-mean-square error (RMSE) for each scenario after optimization. For scenario~$\#$1-4 and the multi-room scenario, the average standard deviation of the optimized positioning RMSEs are $0.17$~cm and $0.36$~cm, respectively. The millimeter-level dispersion indicates the good convergence performance of the proposed algorithms (Algorithm 1 and 2).}
  \label{fig:rebuttal_boxplot}
\end{figure}
We summarize the optimized positioning RMSE of the five scenarios in Figure~\ref{fig:rebuttal_boxplot}. The average standard deviation in scenario~$\#$1-4 is $0.17$~cm. In scenario $\#4$, the optimization results show that the largest standard deviation is ($0.37$~cm). Similarly, the standard deviation in the multi-room scenario is $0.36$~cm. However, the millimeter-level dispersion is insignificant for a centimeter-level accurate UWB-based localization system.

With these experimental results, we can conclude that the optimized anchor placements with the proposed algorithms can provide similar position accuracy with random initialization. Also, the number of required anchor pairs computed in Algorithm~2 does not show a dependency on initial placements. Moreover, with enough computational resources, users can generate multiple initial conditions for optimization and select the sensor placement which gives the best position accuracy.

\end{document}